\pdfoutput=1
\pdfoutput=1
\documentclass[acmtog]{acmart}
\acmSubmissionID{1405}
\AtBeginDocument{%
  \providecommand\BibTeX{{%
    \normalfont B\kern-0.5em{\scshape i\kern-0.25em b}\kern-0.8em\TeX}}}

\definecolor{MyRed}{rgb}{0.65,0.07,0.09}

\definecolor{MyGreen}{rgb}{0.18,0.55,0.09}

\usepackage{wrapfig}
\usepackage{diagbox}
\usepackage{verbatim}
\usepackage{amssymb}
\usepackage{latexsym}
\usepackage{xspace}
\usepackage{color}
\usepackage{natbib}
\usepackage{booktabs}
\graphicspath{{figures/}}
\usepackage{amsmath,bm}
\usepackage{psfrag}
\usepackage{mathtools}

\theoremstyle{definition}
\newtheorem{assumption}{Assumption}
\newtheorem{definition}[assumption]{Definition}
\newtheorem{theorem}{Theorem}
\newtheorem{lemma}[theorem]{Lemma}
\newtheorem{challenge}{Challenge}

\usepackage{booktabs}
\usepackage{subcaption}
\usepackage{graphicx} 
\usepackage{multirow, tabularx}
\newcolumntype{Y}{>{\centering\arraybackslash}X}
\usepackage{capt-of}%
\usepackage{array, boldline, makecell, booktabs}

\usepackage{array}
\usepackage{pifont}
\usepackage{psfrag}
\usepackage{makecell}
\usepackage{algorithm}
\usepackage{threeparttable,booktabs}
\usepackage{graphicx}
\usepackage{subcaption}
\usepackage{flushend}
\usepackage{stfloats} 
\usepackage{appendix}
\usepackage[noend]{algpseudocode}

\usepackage{xcolor}
\usepackage{graphicx}

\newif\ifverbose
\verbosetrue

\ifverbose

\newcommand{\vb}[1]{\textcolor{red}{#1}}
\else

\newcommand{\vb}[1]{}
\fi

\usepackage{url}
\usepackage{xcolor}
\definecolor{newcolor}{rgb}{.8,.349,.1}

\newcommand{\revise}[1]{{#1}}

\begin{document}


\title{A Few-Step Generative Model on Cumulative Flow Maps}


\author{Zhiqi Li}
\authornote{Joint-first author.}
\email{zli3167@gatech.edu}
\orcid{}
\affiliation{%
  \institution{Georgia Institute of Technology}
  \city{Atlanta}
  \country{United States of America}
}

\author{Duowen Chen}
\authornotemark[1]
\email{dchen322@gatech.edu}
\orcid{}
\affiliation{%
  \institution{Georgia Institute of Technology}
  \city{Atlanta}
  \country{United States of America}
}

\author{Yuchen Sun}
\authornotemark[1]
\email{yuchen.sun.eecs@gmail.com}
\orcid{}
\affiliation{%
  \institution{Georgia Institute of Technology}
  \city{Atlanta}
  \country{United States of America}
}

\author{Bo Zhu}
\email{bo.zhu@gatech.edu}
\orcid{}
\affiliation{%
  \institution{Georgia Institute of Technology}
  \city{Atlanta}
  \country{United States of America}
}

\renewcommand{\shortauthors}{Li et al.}
\begin{abstract}

We propose a unified, few-step generative modeling framework based on \emph{cumulative flow maps} for long-range transport in probability space, inspired by flow-map techniques for physical transport and dynamics. At its core is a cumulative-flow abstraction that connects local, instantaneous updates with finite-time transport, enabling generative models to reason about global state transitions.  This perspective yields a unified few-step framework built on cumulative transport and \revise{cumulative} parameterization that applies broadly to existing diffusion- and flow-based models without being tied to a specific prediction \revise{instantiation}.  Our formulation supports few-step and even one-step generation while preserving synthesis quality, requiring only minimal changes to time embeddings and training objectives, and no increase in model capacity. We demonstrate its effectiveness across diverse tasks, including image generation, geometric distribution modeling, joint prediction, and SDF generation, with reduced inference cost.

\end{abstract}


\makeatletter
\@ifundefined{setcctype}{%
  \setcopyright{iw3c2w3}%
  \gdef\@copyrightowner{Copyright is held by the owner/author(s).}%
  \gdef\@copyrightpermission{This work is licensed under a Creative Commons
    Attribution International 4.0 License.}%
}{%
  \setcopyright{cc}%
  \setcctype{by}%
}
\makeatother
\acmJournal{TOG}
\acmYear{2026} \acmVolume{45} \acmNumber{4} \acmArticle{}
\acmMonth{7} \acmDOI{10.1145/3811380}

\begin{teaserfigure}
\centering
\includegraphics[width=\linewidth]{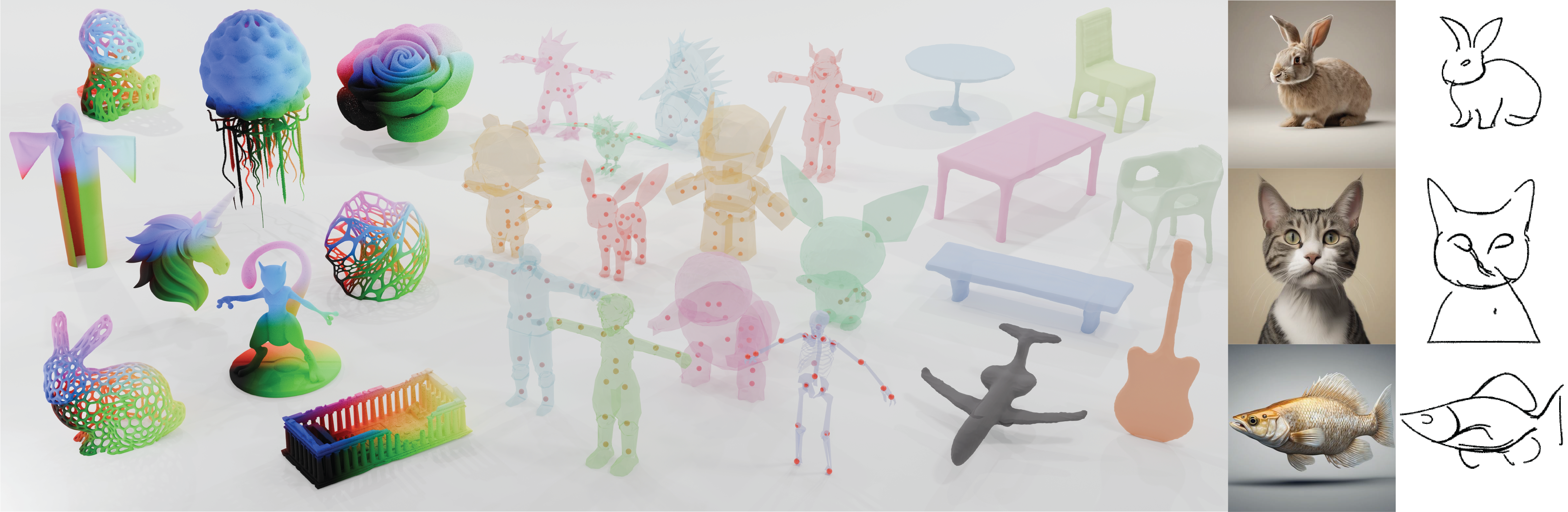}
\caption{\revise{We introduce \emph{Cumulative Flow Maps (CFM)} for few-step generation, a simple training-objective modification that can be incorporated into diverse graphics applications and generative models to accelerate inference without changing the model architecture or using distillation. From left to right, we show geometric distribution modeling with EDM in 6 steps, joint prediction with DDIM in 10 steps, SDF generation sparsely conditioned on 64 surface points with \(x_1\)-FM in 4 steps, and image-conditioned sketch generation with DDIM in 1 step.}}
\label{fig:teaser}
\end{teaserfigure}

\maketitle
\section{Introduction}


Generative models such as \cite{lipman2022flow,song2020denoising,karras2022elucidating} have received increasing attention in computer graphics over the past years, with applications spanning image and video synthesis~\cite{ho2020denoising,geng2024consistency,lipman2022flow,ho2022video}, geometric representations~\cite{zhang2025geometry}, point cloud generation~\cite{wang2025pdt,spadaro2025denoising,mo2023dit}, and implicit field modeling~\cite{zhang2024functional,zhang20233dshape2vecset}. From a flow dynamics perspective, a generative model can be comprehended as learning a \emph{cumulative flow map} \footnote{
\revise{We call the long-range dynamics a \emph{cumulative flow map} because it represents the finite-time accumulation of instantaneous dynamics. As discussed in \autoref{sec:mean_field}, this view naturally extends instantaneous-dynamics parameterizations to long-range transport, yielding a unified long-range parameterization.}} defined in probability space that transports samples from a simple source distribution to a complex data distribution (e.g., flow matching ~\cite{lipman2024flow} and flow map matching~\cite{boffi2025flow}). This cumulative flow map represents a finite-time, global transport that directly maps an initial state $x_0 \sim p_0$, typically drawn from a standard Gaussian or uniform distribution, to a final state $x_1 \sim p_{\text{data}}$, such that the pushforward of $p_0$ under this map matches the target data distribution. In this view, generative sampling amounts to evaluating the learned cumulative flow map, either explicitly or implicitly, with model quality determined by how accurately the induced transport reproduces the statistics and geometric structure of the data distribution.

In practice, directly learning \emph{cumulative} flow maps is challenging due to the long-horizon and highly nonlinear nature of transport, making network training difficult. As a result, most generative models instead learn \emph{instantaneous flow maps} (\revise{or instantaneous dynamics}, e.g., velocity fields in flow matching~\cite{dao2023flow,lipman2022flow}) that predict local state updates conditioned on the current state $x(t)$, which are then composed through numerical integration to approximate the desired cumulative transport. To reduce reliance on multi-step integration, recent work has explored few-step and one-step generation (e.g.,~\cite{frans2025shortcut,zhou2025inductive,geng2025mean}). In particular, Mean Flow~\cite{geng2025mean} approximates finite-time transport by learning a \emph{mean velocity} along the trajectory, reformulating the objective so that the cumulative effect can be captured in a single update. While effective for image generation, this approach is intrinsically tied to the $u$-prediction flow-matching formulation and does not naturally generalize to other widely used frameworks such as DDIM~\cite{song2020denoising}, EDM~\cite{karras2022elucidating}, or $x_1$-prediction flow matching~\cite{lipman2022flow}, which are central to graphics and visual computing applications.

To address these challenges in modeling long-range, cumulative flow maps for generative tasks, we propose a unified abstraction that connects local state transitions to long-range transport via an explicit cumulative-field parameterization. This formulation provides a principled bridge between instantaneous updates in existing generative models and the cumulative transport governing data generation. Building on this abstraction, we introduce a unified learning framework, termed \emph{Cumulative Flow Maps (CFM)}, that generalizes Mean Flow beyond the $u$-prediction setting and applies across a range of generative formulations, including $u$- and $x_1$-flow matching, EDM, and DDIM, thereby enabling few-step and even one-step generation in settings where $u$-prediction methods are not applicable, such as geometry distribution~\cite{zhang2025geometry}, SDF generation~\cite{zhang2024functional}, and pixel-space image generation~\cite{li2025back}. \revise{This flexibility highlights CFM as a general mathematical framework particularly suitable for graphics applications involving diverse data representations and generative formulations. In particular, Mean Flow can be viewed as a special instantiation of CFM under the $u$-prediction flow-matching formulation.}

To enable practical training of cumulative flow maps within existing generative pipelines, we derive a field-equation-based formulation that transforms the cumulative parameterization and introduces principled supervision from the data distribution. This unified approach enables self-consistent learning of long-range transport fields across diverse generative frameworks despite the lack of conditional structure in standard objectives. The method is model-agnostic and requires no architectural changes or distillation; modifying only the training objective substantially reduces sampling steps. In practice, it achieves $10\times$--$200\times$ reductions in inference-time cost while preserving, and often improving, generation quality across graphics generative tasks. It is worth mentioning that the cumulative flow map design is motivated by the concurrent flow-map techniques developed in fluid simulation, where long-range transport and global state evolution are modeled directly in physical space without explicit mean-velocity parameterization~\cite{deng2023fluid,junwei2024,li2024lagrangian,li2025clebsch,li2025edge}.


Our contributions can be summarized as follows:
\begin{itemize}
    \item \textbf{Instantaneous--cumulative flow-map abstraction.} We formalize cumulative flow maps as finite-time transport obtained by composing instantaneous flow maps, unifying multi-step and few-step generation.
    \item \textbf{Cumulative flow map learning beyond $u$-prediction.} We introduce a unified framework that generalizes Mean Flow to $u$- and $x_1$-flow matching, EDM, and DDIM for graphics generation.
    \item \textbf{Model-agnostic training via field equations.} We derive a field-equation-based objective that learns cumulative flow maps without architectural changes, substantially reducing sampling steps in existing models.
\end{itemize}

\begin{figure}[t]
    \centering
    \includegraphics[width=1.3\linewidth]{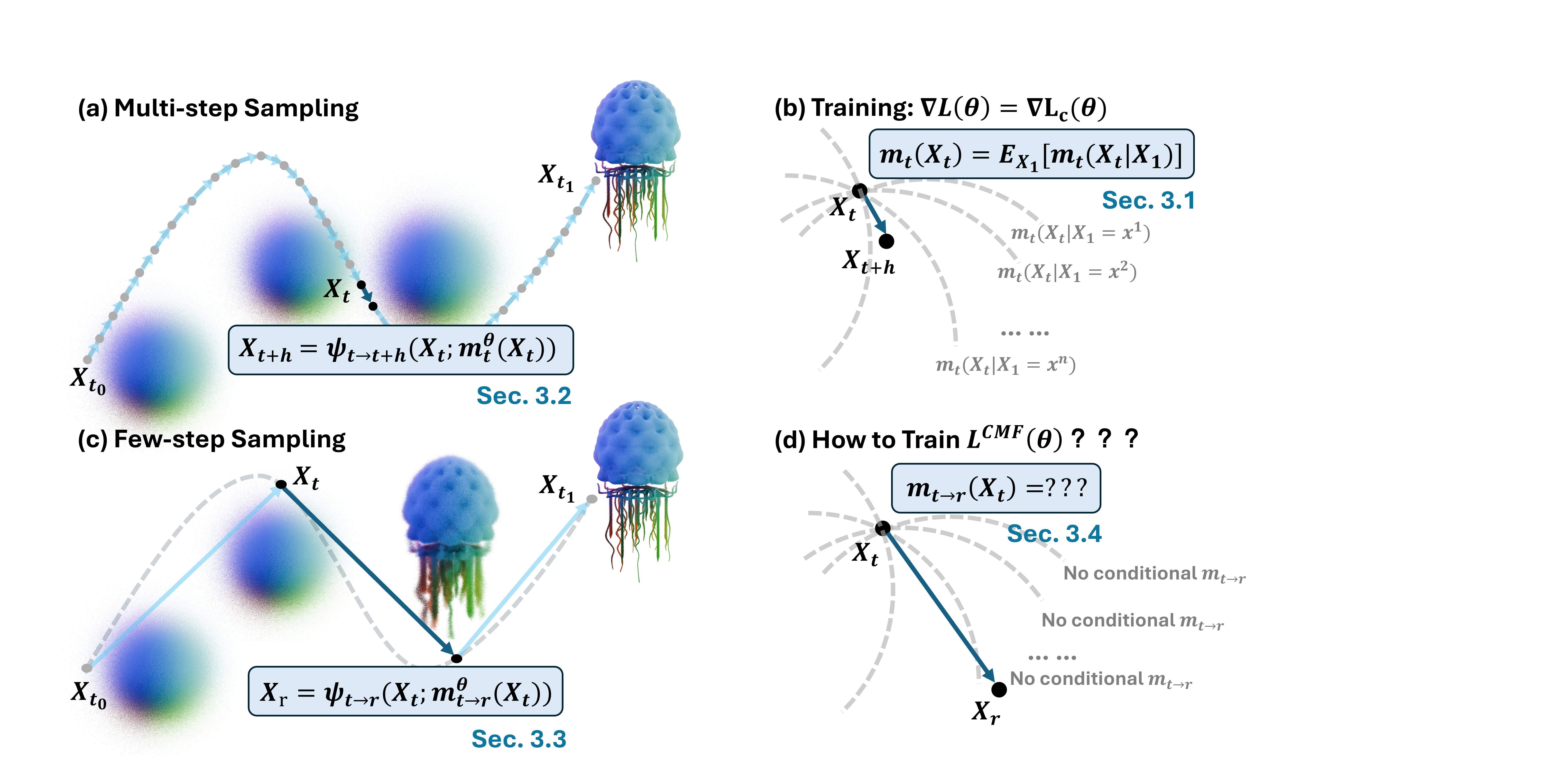}
    \vspace{-3mm}
    \caption{Illustration of multi-step and few-step generation.
    (a,b) Multi-step generation sampling and training.
    Sampling proceeds via instantaneous flow maps $\psi_{t\to t+h}(X_t; m_t^\theta)$ with a small step size $h$.
    The instantaneous model $m_t^\theta$ is trained using its conditional counterpart $m_t(X_t \mid X_1)$, where dashed curves in (b) indicate conditional paths.
    (c,d) Few-step generation sampling and training.
    Sampling uses long-range flow maps $\psi_{t\to r}(X_t; m_{t\to r}^\theta)$ that advance directly from time $t$ to an arbitrary future time $r$.
    The parametrized long-range models $m_{t\to r}^\theta$ are referred to as \emph{mean fields}.
    \textbf{However, these mean fields admit no conditional counterpart, which poses a challenge for training; our solution is discussed in \autoref{sec:training_meanfield}}.
    }
    \vspace{-3mm}
    \label{fig:illustration}
\end{figure}

\section{Related Work}
\paragraph{Few-Step Generation}
Reducing sampling steps is a central challenge for improving the efficiency of diffusion and flow-based generative models. A dominant line of work addresses this problem through distillation, enabling few-step or even one-step generation. Such approaches have been extensively explored for diffusion models \cite{salimans2022progressive, meng2023distillation, geng2023onestep, axel2024adver, luo2024diff, yin2024one, zhou2024score} and later extended to flow-based formulations \cite{liu2023flow}. As an alternative to teacher–student distillation, consistency models\cite{song2023consistency, song2023improved, lu2025simplifying} were introduced as independently trainable one-step generators. Inspired by this paradigm, recent studies have incorporated self-consistency principles into broader generative frameworks, including Shortcut Models \cite{frans2025shortcut} and multi-step stochastic interpolation schemes \cite{zhou2025inductive}. Mean Flow \cite{geng2025mean, geng2025improved} models time-averaged velocities through differentiation of the Mean Flow identity, achieving state-of-the-art performance for one-step generation on ImageNet. However, these methods are primarily designed for image generation, and their complex formulations make them difficult to adapt or translate directly to graphics applications.

\begin{figure*}[t]
    \includegraphics[width=1.0\textwidth]{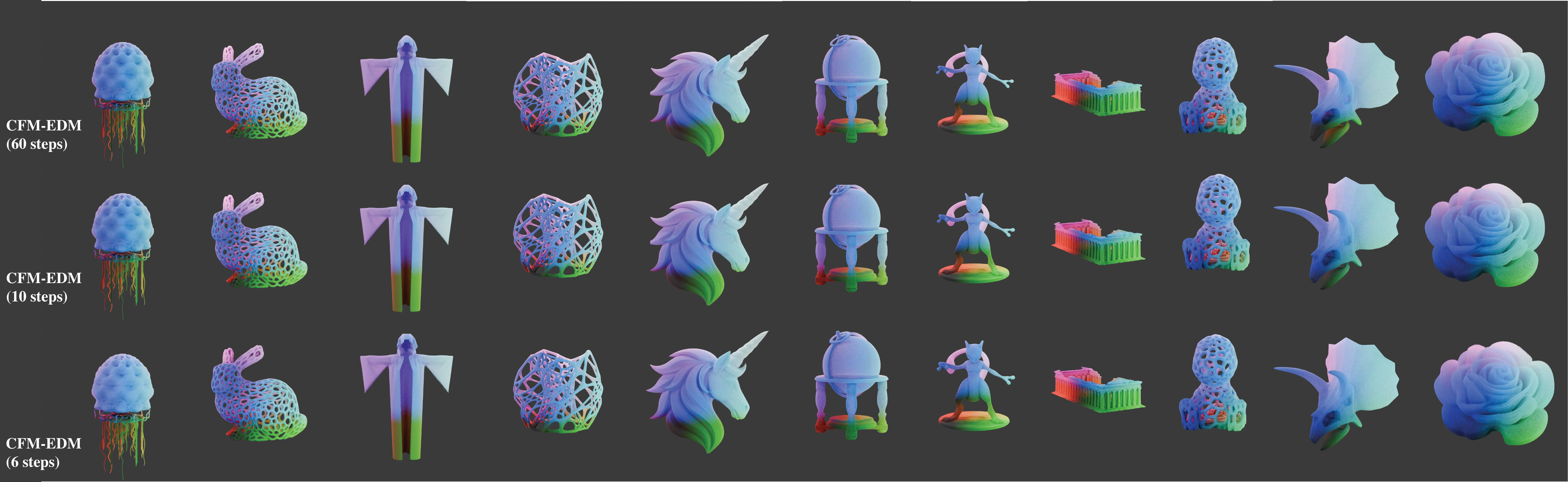}
    \vspace{-5mm}
    \caption{\revise{Geometric distribution generated using our CFM-EDM method. Our method achieves a \(10\times\) speedup without degrading generation quality. Notably, it only modifies the training loss, without changing the network architecture or relying on distillation.}}
    \vspace{-3mm}
    \label{fig:geodist_grid}
\end{figure*}

\paragraph{Flow Map Methods}
The concept of the Flow Map originates from differential geometry and dynamical systems \cite{arnold1992ordinary}, describing the evolution of points under a time-dependent vector field.  Flow maps were initially developed in geophysical and hydrological modeling and have since been widely adopted in geometric modeling \cite{desbrun2006discrete} and fluid simulation \cite{nabizadeh2022covector,tessendorf2011characteristic}.  Both fluid simulation and multi-step generative models face a common challenge: reliance on iterative time integration leads to high computational cost and error accumulation. 
To address this issue, a variety of flow-map–based methods have been proposed in fluid simulation, including grid-based\cite{li2025edge,sun2024igfm}, particle-based\cite{li2024lagrangian}, neural\cite{deng2023fluid}, and hybrid approaches \cite{li2024particle,chen2024solid}, which improve efficiency by directly modeling long-range transport.  More recently, flow maps have been explored for accelerating generative models. Flow Map Matching~\cite{boffi2025flow} enables few-step generation by directly learning long-range mappings or \revise{using $u$-prediction parameterization, but such direct regression or $u$-prediction parameterization can be unstable and tends to degrade generation quality in some graphics applications. In contrast, our work leverages flow-map principles to connect instantaneous and long-range flow maps and provides a unified long-range parameterization that naturally extends various instantaneous-dynamics parameterizations}, enabling stable learning and naturally supporting few-step and one-step generation.

\paragraph{Generative Method for Graphics}
Generative methods are widely studied in computer graphics, spanning animation~\cite{li2025dress, huang2025animax, ghosh2025duetgen}, geometry~\cite{wei2025octgpt, lai2025hunyuan3d25highfidelity3d, ye2025hi3dgen}, rendering~\cite{gu2025diffusion, zeng2025renderformer, chen2025fluid}, and reconstruction~\cite{yao2025cast, wang2025vggt, liao2025pad3r}. We focus on three related areas: (1) point cloud generation, (2) joint location prediction, and (3) implicit geometry representations. Early point cloud models used normalizing flows~\cite{yang2019pointflow} or VAEs~\cite{achlioptas2018learning}, while diffusion-based methods~\citep{luo2021diffusion, zhou20213d} further improved quality (e.g., LION~\citep{vahdat2022lion}, ShapeGF~\cite{cai2020learning}, TIGER~\cite{ren2024tiger}) and recent work incorporates level-of-detail strategies~\cite{meng2025pointnsp}. Auto-rigging has evolved from classical methods~\cite{baran2007automatic} to data-driven joint prediction using volumetric networks~\cite{xu2019predicting}, graph networks~\cite{RigNet}, and diffusion models~\cite{wang2025pdt}; our approach is closest to PDT~\cite{wang2025pdt} while offering up to 
200$\times$ faster inference. Finally, implicit representations compress geometry via neural fields~\cite{park2019deepsdf, mescheder2019occupancy, sitzmann2020implicit} with acceleration from auxiliary structures~\cite{takikawa2021neural, muller2022instant}, and recent diffusion-based distribution matching further strengthens point-cloud-based modeling~\cite{zhang2025geometry}.

\section{Method}
In this section, we introduce multi-step generative models from the perspective of instantaneous flow maps (\autoref{sec:fundamental}) and abstract them into a unified framework (\autoref{sec:abstraction}). We then extend instantaneous flow maps to cumulative flow maps and define their cumulative parameterization for one-step and few-step generation (\autoref{sec:mean_field}), followed by a discussion of the challenges and solutions for training the cumulative parameterization. (\autoref{sec:training_meanfield}).

\subsection{Instantaneous Flow Maps for Multi-Step Generation}\label{sec:fundamental}
\paragraph{Continous-Time Markov Process} Score Matching \cite{song2020score}, Diffusion Models \cite{song2020denoising, ho2020denoising,karras2022elucidating}, and Flow Matching \cite{dao2023flow, lipman2022flow} methods can be unified under a continuous-time Markov process (CTMP).  We consider a stochastic process $\{X_t\}_{t \in \mathcal{I}}$ on a state space $\mathcal{X}$, parameterized by an evolution parameter $t$ from an initial value $t_{0}$ to a terminal value $t_{1}$. The process satisfies the Markov property and is characterized by its short-time transition kernel $p_{t+h\mid t}(A \mid x) := P[X_{t+h}\in A \mid X_t = x], \forall A \subset \mathcal{X}$.  A generative model learns a time-dependent function $m_t(x)$ (e.g. $v_t(x)$ for Flow Matching), which parameterizes the local transition behavior via $p_{t+h\mid t}(A \mid x) = p_{t+h\mid t}(A \mid x; m_t(x)) + O(h)$.
Generation is performed by forward sampling: starting from $X_{t_0} \sim p_{t_0}$ with a given distribution $p_{t_0}$, samples are propagated by repeatedly drawing $X_{t+h} \sim p_{t+h\mid t}(\cdot \mid X_t; m_t)$ until $t=t_1$, yielding a generated sample $X_{t_1}$.

\paragraph{Training CTMP} For training $m_t^\theta(x)$ with the direct objective $\mathcal L(\theta)=\mathbb E_{t,x\sim p_t}\|m_t^\theta(x)-m_t(x)\|_2^2$, where $\theta$ denotes the model parameters and $m_t(x)$ is the reference function, the main challenge is that both the reference $m_t(x)$ and the marginal distribution $p_{t}$ are unknown in practice.  To address this issue, generative models construct a conditional distribution $p_t(\cdot|X_{t_1})$ with an explicit analytical form defined with respect to $X_{t_1} \sim p_{\mathrm{data}}=p_{t_1}$, and marginalizing the corresponding conditional quantities can obtain the marginal distribution $P_t(x) = E_{X_1\sim p_{data}}[P_t(x|X_{t_1})]$, transition kernel $p_{t+h|t}(A|x) = E_{X_{t_1}\sim p_{data}}[p_{t+h|t}(A|x,X_{t_1})]$, and parameterized function $m_t(x)= E_{X_{t_1}\sim p_{data}}[m_t(x|X_{t_1})]$.  Using the conditional distribution $P_t(x|X_{t_1})$ and the conditional parameterized function $m_t(x|X_{t_1})$, a surrogate objective $\mathcal L_c(\theta) = \mathbb E_{t,x\sim P_t(\cdot|x_{t_1}), x_{t_1}\sim P_{data}}\|m_t^\theta(x) - m_t(x|X_{t_1} = x_{t_1})\|_2^2$ is constructed, and generative models show that this surrogate objective satisfies $\nabla_\theta \mathcal L_c(\theta) = \nabla_\theta \mathcal L(\theta)$ \cite{song2020score,lipman2022flow,ho2020denoising,song2020denoising}, allowing it to be used to learn $m_t^\theta(x)$ for the target objective $\mathcal L(\theta)$.

\begin{figure*}[t]
\includegraphics[width=1.0\textwidth]{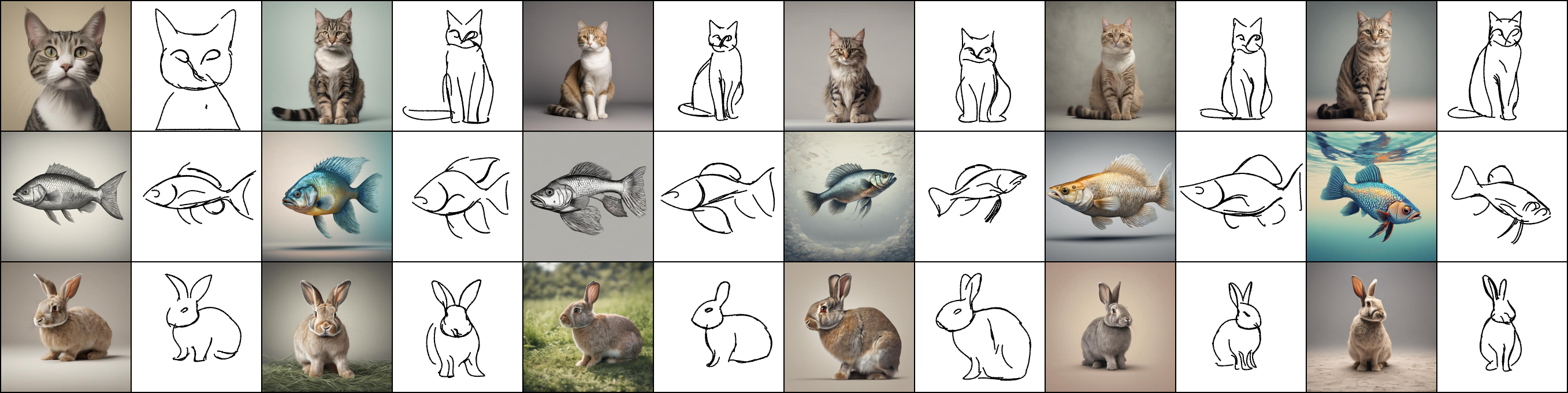}
\vspace{-5mm}
\caption{\revise{Sketch generation results on unseen images. 1-step CFM-DDIM achieves visual fidelity comparable to the prior 50-step diffusion-based method~\cite{arar2025swift}, while requiring only one sampling step. This improvement is achieved by modifying only the training loss, without changing the model architecture.}}
\vspace{-3mm}
\label{fig:sketch_grid}
\end{figure*}

\paragraph{Sampling with \textbf{Instantaneous Flow Maps}} Many generative models under the same distribution path $(P_t)_{0 \le t \le 1}$, can be described using deterministic transitions \cite{lipman2022flow,song2020denoising,song2020score}. Specifically, for sufficiently small $h$, there exists a deterministic function $\psi_{t\to t+h}:S\to S$ such that $p_{t+h\mid t}(\delta_{\psi_{t\to t+h}(x)} | x)=1$,  where $\delta_{\psi_{t\to t+h}(x)}$ denotes a point mass at $\psi_{t\to t+h}(x)$, while preserving the distribution path $(P_t)_{t\in \mathcal{I}}$.  We refer to $\psi_{t \to t+h}: S \to S$ as \textbf{instantaneous flow maps}, as they determine how a state at time $t$ is deterministically mapped to a nearby future time $t+h$.  Since sampling can be realized through such deterministic transitions, which typically yield higher sampling quality and deterministic behavior \cite{song2020denoising}, we focus on deterministic transitions here.  $u-$FM, $x_1-$FM, DDIM, and EDM are instantiated as follows:

\begin{enumerate}
    \item \textbf{\revise{$u-$FM:}} $t_0 \!=\! 0$, $t_1 \!=\! 1$; $m(t) \!=\! u_t(x)$; $\psi_{t\to t+h}(x) \!=\!x + u_t(x)h+O(h)$; $P_t(X|X_{t_1}\!=\!x_{t_1})\!=\!\mathcal{N}(tx_{t_1}, (1-t)^2I)$; $u_t(x|x_{t_1})\!=\!\frac{x_{t_1}-x}{1-t}$.
    \item \textbf{\revise{$x_1-$FM:}} $t_0 \!=\! 0$, $t_1 \!=\! 1$; $m(t) \!=\! x^1_t(x)$; $\psi_{t\to t+h}(x) \!= \!x + \frac{x^1_t(x)-x}{1-t}h+O(h)$; $P_t(X|X_{t_1}\!=\!x_{t_1})\!=\!\mathcal{N}(tx_{t_1}, (1-t)^2I)$; $x^1_t(x|x_{t_1})=x_{t_1}$ .
    \item \textbf{\revise{DDIM:}} $t_0 \!=\! T$, $t_1 \!=\! 0$; $m(t) \!=\! \tilde x_t(x)$;  $\psi_{t\to t+h}(x) \!=\! \sqrt{\bar\alpha_{t+h}}\tilde x_{t}(x_t) + \sqrt{1-\bar\alpha_{t+h}}\frac{x_t - \sqrt{\bar\alpha_t}\tilde x_{t}(x_t)}{\sqrt{1-\bar\alpha_t}}$ + O(h); $P_t(X|X_{t_1}\!=\!x_{t_1})\!=\!\mathcal{N}(\sqrt{\bar\alpha_t}x_{t_1}, (1-\bar\alpha_t)I)$; $\tilde x_t(x|x_{t_1})\!=\!x_{t_1}$. (Here we show the $x$-prediction.). \footnote{
Although DDIM can be described within a score-matching formulation~\cite{song2020score}, the formulation adopted here explicitly captures attraction toward the data manifold~\cite{liu2022pseudo}, which improves numerical stability and makes it the default inference scheme in widely deployed diffusion frameworks (e.g., Stable Diffusion~\cite{vonplaten2022diffusers}). As such, it represents the fundamental formulation.}
    \item \textbf{\revise{EDM:}} $t_0 \!=\! \sigma_{\max}$, $t_1 \!=\! 0$; $m(t) \!=\! D_t(x)$;  $\psi_{t\to t+h}(x) \!=\! x+h\frac{x-D_t(x)}{t}+O(h)$; $P_t(X|X_{t_1}\!=\!x_{t_1})\!=\!\mathcal{N}(x_{t_1}, tI)$; $D_t(x|x_{t_1}) \!=\! x_{t_1}$.
\end{enumerate}

\noindent where $\mathcal{N}(\cdot,\cdot)$ denotes a normal distribution and $I$ denotes the identity matrix.  For DDIM, the noise scheduler is specified on the discrete time steps $t = T,\ldots,0$, with $\alpha_{t} = 1 - \beta_{t}$, $\bar\alpha_{t} = \prod_{s=0}^{t} \alpha_{s}$ and $\beta_{t} = \beta_0 + \frac{t}{T}(\beta_T - \beta_0)$ ($\beta_0$ and $\beta_T$ are fixed constants).


\subsection{Unified Representation of Instantaneous Flow Maps}
\begin{figure}[t]
    \centering
    \includegraphics[width=\linewidth]{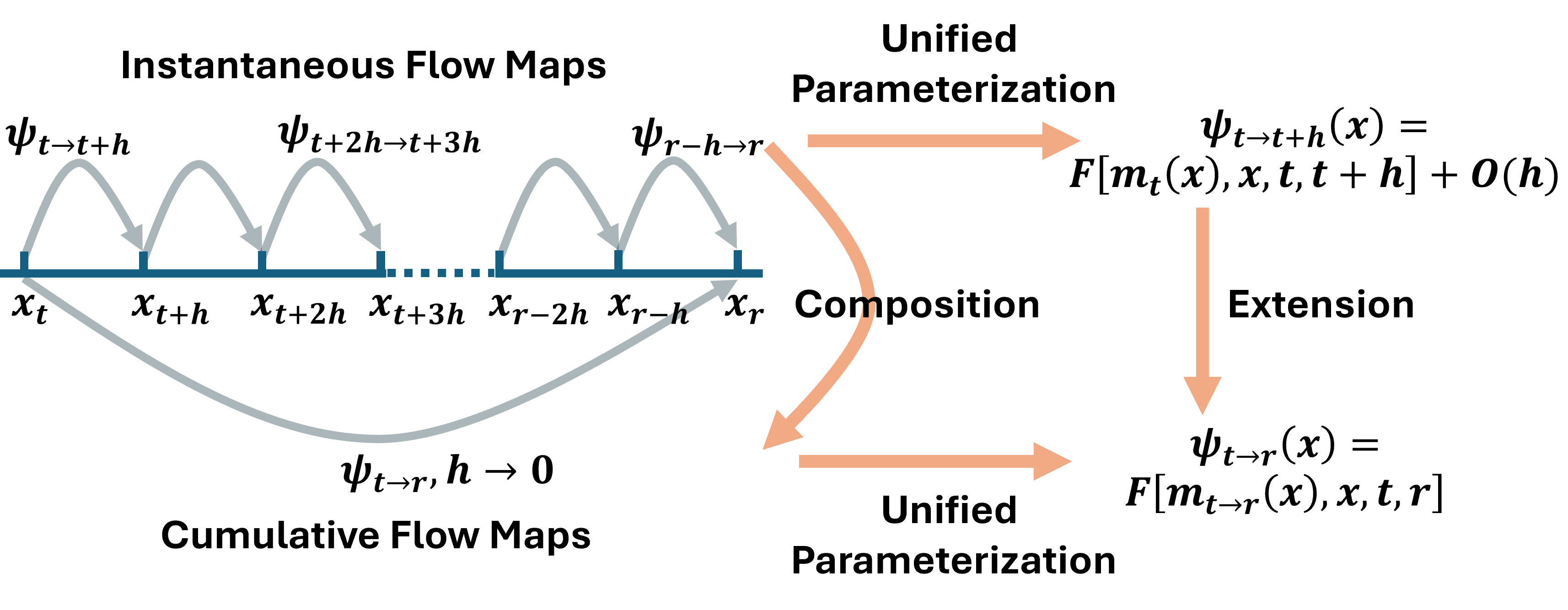}
    \vspace{-10mm}
    \caption{Relationships among flow map concepts.  An instantaneous flow map $\psi_{t\to t+h}$ is represented in a unified form using a parametrized field $m_t(x)$ and a abstract function $F[\cdot,\cdot,\cdot,\cdot]$.  By taking the limiting composition of instantaneous flow maps, a cumulative flow map $\psi_{t\to r}$ is defined.  Under the same representation, cumulative parametrization $m_{t\to r}(x)$ are further introduced and defined to satisfy $\psi_{t\to r}(x) = F[m_{t\to r}(x),x,t,r]$, serving as the quantities learned by the model.
}
\vspace{-5mm}
    \label{fig:definition}
\end{figure}

\label{sec:abstraction}
To provide a unified representation of $\psi_{t\to t+h}(x)$, we express it as $\psi_{t\to t+h}(x) = F[m_t(x), x, t, t+h]+O(h)$, where $F[f_1, f_2, f_3, f_4]$ \revise{is an abstract function} that takes four values $f_1, f_2, f_3, f_4$ as inputs. We denote the partial derivative of $F$ with respect to its $i$-th argument by $\partial_i F = \frac{\partial F}{\partial f_i}$, and make the following assumptions on $F$:
\begin{assumption}[Constraining the abstract function $F$]\label{assumpt:constraining_F} We make the following smoothness assumptions on $F$:
\begin{enumerate}
    \item \textbf{(Differentiability)} $F$ is smooth with respect to each $f_i$.
    \item \textbf{(Invertibility)} The mixed partial derivative of $F$ with respect to $f_1,f_4$ and $f_1,f_3$ are invertible a.e., i.e., $\frac{\partial^2 F}{\partial f_1\partial f_4}$ and $\frac{\partial^2 F}{\partial f_1\partial f_3}$ are non-degenerate a.e.
    \item \textbf{(Identity)} When $f_3 = f_4$, we have $F[f_1, f_2, f_3, f_4] = f_2$.
    \item \textbf{(Affine structure)} There exist abstract functions $P[
\cdot,\cdot]$ and $Q[\cdot,\cdot]$ such that $F$ can be written as $F[f_1,f_2,f_3,f_4] = P[f_3,f_4]f_1 + Q[f_3,f_4]f_2$. 
\end{enumerate}

\end{assumption}
\begin{figure*}[t]
\includegraphics[width=1.0\textwidth]{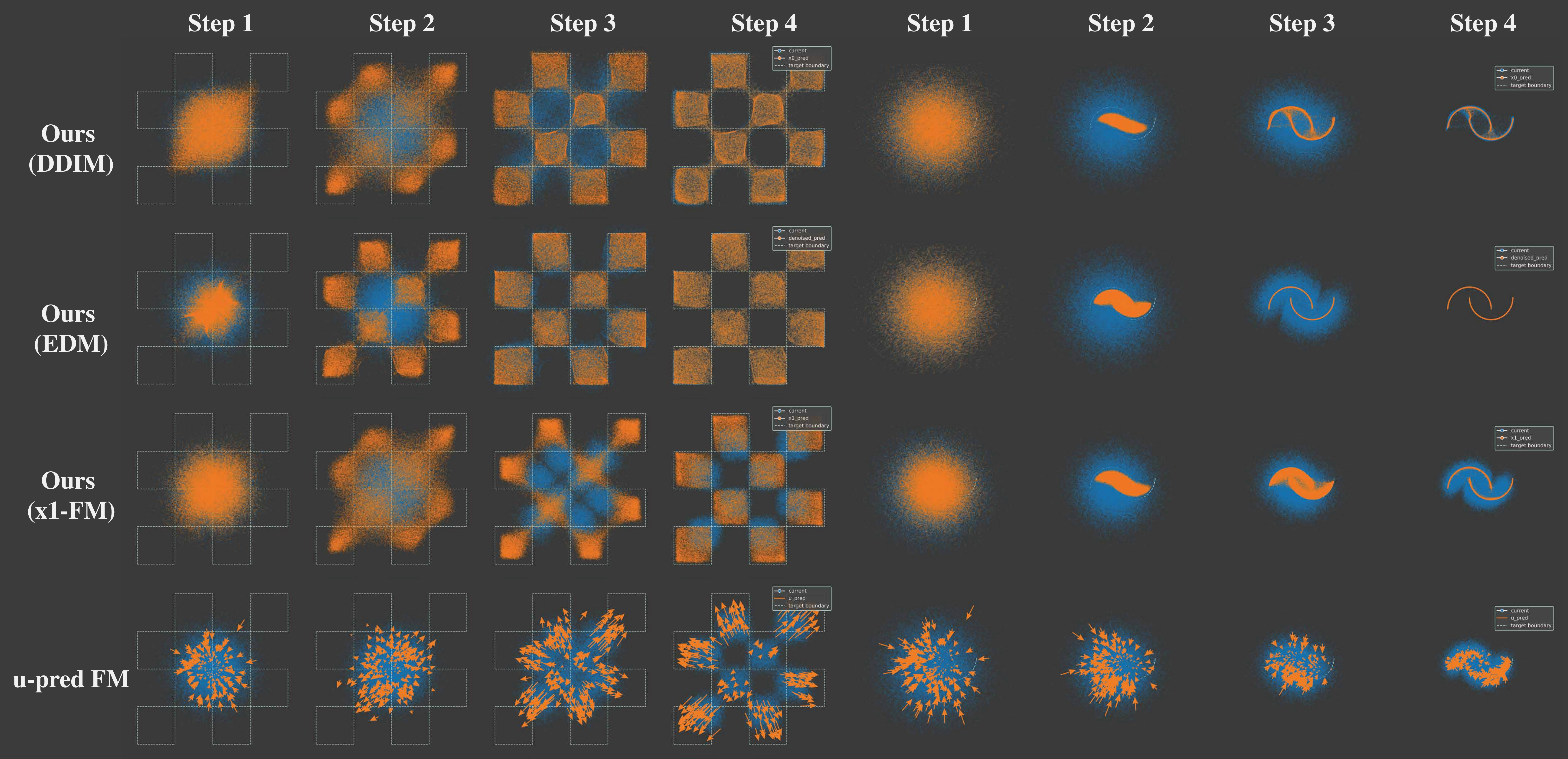}
\caption{\revise{Toy examples on the Checkerboard and Two-Moons datasets. For 4-step CFM-based generation, we show, at each step, the intermediate sample locations (blue) together with the corresponding predicted targets (orange) for different formulations: DDIM, $x_1$-FM, EDM, and $u$-FM (equivalent to MeanFlow). This visualization illustrates how both the sample trajectory and the prediction target evolve across steps under different parameterizations.}}
\label{fig:toy_example}
\end{figure*}
\revise{Here, the differentiability and invertibility assumptions ensure that the induced dynamics are locally well defined and smoothly varying. The identity condition guarantees zero-gap consistency, so that the flow map reduces to the current state when the start and end times coincide. The affine-structure assumption reflects the form shared by common generative parameterizations.} We adopt the abstract function $F$ to decouple variable dependencies and provide a unified representation of $\psi_{t\to t+h}$ across different parameterizations.
\begin{enumerate}
    \item \textbf{\revise{$u$-FM:}} $F[f_1,f_2,f_3,f_4] = f_1(f_4-f_3)+f_2$, $f_1=u_t(x)$
    \item \textbf{\revise{$x_1$-FM:}} $F[f_1,f_2,f_3,f_4] =\frac{f_1-f_2}{1-f_3}(f_4-f_3) + f_2$, $f_1=x^1_t(x)$
    \item \textbf{\revise{DDIM:}} $F[f_1,f_2,f_3,f_4] \!=\! \sqrt{\bar\alpha_{f_4}}f_1 \!+\! \sqrt{1-\bar\alpha_{f_4}}
    \frac{f_2 \!-\! \sqrt{\bar\alpha_{f_3}}f_1}
    {\sqrt{1-\bar\alpha_{f_3}}}$, $f_1\!=\!\tilde{x}_t(x)$
    \item \textbf{\revise{EDM:}}  $F[f_1,f_2,f_3,f_4] = (f_4-f_3)\frac{f_2-f_1}{f_3}+f_2, f_1=D_t(x)$
\end{enumerate}

\noindent where $f_2 = x$, $f_3 = t$ and $f_4 = t+h$. All these instantiations of $F$ can be readily verified to satisfy \revise{Assumption} \autoref{assumpt:constraining_F}.

From the formulation above, we observe that the instantaneous
flow map $\psi_{t\to t+h}(x) = F[m_t(x), x, t, t+h] + O(h)$ depends on the state $x$ and the function $m_t(x)$ at time $t$, and therefore can only advance the sampling process from $t$ to a nearby time $t+h$.  As a result, the generative process requires repeatedly applying such instantaneous
flow maps, starting from noise at $t=t_0$ and progressing through many steps until reaching the data distribution at $t=t_1$ (e.g., using $h=\frac{t_1-t_0}{1000}$ results in $1000$ iterative steps).  \textbf{This multi-step generation incurs substantial computational cost.
Building on the unified representation $F[m_t(x), x, t, t+h]$, we extend instantaneous flow maps to long-range cumulative flow maps and parameterize them using cumulative fields $m_{t\to r}(x)$.
This leads to a unified few-step (including one-step) generative framework, which we refer to as \emph{Cumulative Flow Maps (CFM)}, substantially reducing the number of sampling steps without sacrificing generation quality.}

\subsection{Cumulative Flow Maps for Few-Step Generation} \label{sec:mean_field}

The core idea of Mean Flow \cite{geng2025mean} is to replace the learning of the instantaneous velocity $u_t(x)$ in $u$-prediction Flow Matching with the learning of an average velocity $u_{t\to r}(x) = \frac{\psi_{t\to r}(x) - x}{r - t}$, where $\psi_{t\to r}(x) = \int_t^r u_\tau(\psi_{t\to \tau}(x)) d\tau $ is the natural long-range extension of the local transition $\psi_{t\to t+h}(x)$.  This formulation enables both one-step $x_1 = u^\theta_{0\to 1}(x_0) + x_0$ and few-step generation, for example, two-step generation $x_{0.5} = x_0 + 0.5u^\theta_{0\to 0.5}(x_0)$ and $x_1 = x_{0.5} + 0.5u^\theta_{0.5\to 1}(x_{0.5})$.  To extend this idea beyond Flow Matching to other generative frameworks, we leverage our unified formulation based on $F[\cdot,\cdot,\cdot,\cdot]$ and define the notion of cumulative flow maps and their cumulative parameterization fields in a general setting.

\begin{definition}[Cumulative Flow Maps]\label{def:mean_fields} Given the instantaneous flow map $\psi_{t\to t+h}$ and the abstract function $F[\cdot,\cdot,\cdot,\cdot]$, the natural long-range cumulative extension of the flow map $\psi_{t\to r}$ is defined as  
\begin{equation}
    \begin{aligned}
        \psi_{t\to r}(x) = \lim_{\max_{i} \{t_i-t_{i-1}\}\to \infty}\psi_{t_{n-1}\to r}(\psi_{t_{n-2}\to t_{n-1}}(....\psi_{t\to t_1}(x)))   
    \end{aligned}
\end{equation}
for any partition $\{t_i\}_{i=0}^n$ of the interval $[0,1]$ with $0 = t_0 < t_1 <...< t_n = 1$.  The cumulative flow maps $\psi$ satisfies the semigroup property: for any $t < s < r$, $\psi_{t\to r}(x) = \psi_{s\to r}(\psi_{t\to s}(x))$.  We then define the cumulative parameterization field $m_{t\to r}$ as the field satisfying
\begin{equation}
    \begin{aligned}
  \psi_{t\to r}(x) = F[m_{t\to r}(x), x, t, r].      
    \end{aligned}
\end{equation}
\end{definition}
The cumulative parameterization field $m_{t\to r}(x)$ supports both one-step and few-step generation (Algorithm~\autoref{alg:sampling}), and its consistency with the instantaneous field, as shown in the following property, stabilizes Algorithm~\autoref{alg:training} and also allows $m_{t\to r}^\theta$ to be learned from the pre-trained instantaneous field $m_t^\theta$, substantially accelerating training (see \autoref{tab:ablation_50k_multi}).
\begin{theorem}[Consistency Between the Cumulative Field and the Instantaneous Field]\label{thm:limitation}
The cumulative parameterization $m_{t\to r}(x)$ defined in Definition~\ref{def:mean_fields} satisfies
\begin{equation}
    \begin{aligned}
        \lim_{r \to t} m_{t\to r}(x) = m_t(x).
    \end{aligned}
\end{equation}
See Supplement A.1 for a proof.
\end{theorem}

\subsection{Training Cumulative Parameterization} \label{sec:training_meanfield}

To learn $m^\theta_{t\to r}(x)$, a direct objective is to minimize the direct loss $\mathcal{L}^{CMF}(\theta)=\|m^\theta_{t\to r}(x)-m_{t\to r}(x)\|_2^2$; however, since no reference cumulative field $m_{t\to r}(x)$ can be analytically computed from the data distribution, $\mathcal{L}^{CMF}(\theta)$ cannot be used for training from scratch.  A natural idea is to construct a conditional cumulative field $m_{t\to r}(x | X_{t_1})$ with supervision from the dataset, analogous to multi-step generative models, and use it to define a surrogate loss $\| m^\theta_{t\to r}(x) - m_{t\to r}(x | X_1)\|_2^2$,  but the following statement shows that this is impossible and poses a fundamental challenge for training $m_{t\to r}$.

\begin{challenge}[Non-existence of Conditional Cumulative Fields] There exists no conditional cumulative field $m_{t\to r}(x \mid X_{t_1})$ that simultaneously (i) is consistent with the conditional path transition $F[\cdot,\cdot,\cdot,\cdot \mid X_{t_1}]$ under the Definition \ref{def:mean_fields}, and (ii) satisfies the consistency relation $m_{t\to r}(x)=\mathbb{E}_{X_{t_1}\sim p_{data}}[m_{t\to r}(x|X_{t_1})]$ with marginal cumulative fields. As a result, a self-consistent conditional cumulative field does not exist. (See Supplement A.2 for a proof.)
\end{challenge}

To address this challenge, we reformulate the cumulative field $m_{t\to r}(x)$ into an equivalent form that expresses $m_{t\to r}(x)$ in terms of the instantaneous field $m_t(x)$ and the derivatives of $m_{t\to r}(x)$.  After substituting this expression into $\mathcal{L}^{CMF}(\theta)$, we introduce supervision from the dataset by exploiting the conditional $m_{t}(x|X_{t_1})$.

\begin{lemma}[Initial-Time Derivative of the Cumulative Flow Map $\psi_{t\to r}$] The initial-time derivative of the cumulative flow map $\psi_{t\to r}$ defined in Definition \autoref{def:mean_fields} can be expressed as
\begin{equation}
    \begin{aligned}
\partial_t\psi_{t\to r}(x)
=
-(\partial_x\psi_{t\to r}(x))
\revise{[}\partial_\tau\psi_{t\to \tau}(x)\revise{]}|_{\tau=t}
    \end{aligned}
\end{equation}
See Supplement A.3 for a proof.
\end{lemma}
\begin{theorem}[A Reformulation of the Cumulative Field]\label{thm:reformulation}
There exist a sufficiently smooth abstract function $E[\cdot,\cdot,\cdot,\cdot,\cdot,\cdot,\cdot]$, which is affine with respect to its last argument, and scalar-valued abstract functions \(G[\cdot,\cdot]\) and \(H[\cdot,\cdot]\),
satisfying $G[f_3,f_4]|_{f_3=f_4}=1$ and $H[f_3,f_4]|_{f_3=f_4}=0$, such that, for almost every $(x,t,r)$, the cumulative field $m_{t\to r}(x)$ admits the following representation:
\begin{equation}\label{eq:reformulation}
\begin{aligned}
m_{t\to r}(x)
&= G(t,r)\, m_{t\to t}(x)
+ H(t,r)\,
E[
m_{t\to r}(x), x, t, r, \\
&\qquad\qquad
\partial_t m_{t\to r}(x),
\partial_x m_{t\to r}(x),
m_{t\to t}(x)
].
\end{aligned}
\end{equation}
Moreover, within $E[\cdot,\cdot,\cdot,\cdot,\cdot,\cdot,\cdot]$, the dependence on $\partial_t m_{t\to r}(x)$ and $\partial_x m_{t\to r}(x)$ appears only through the combined term
\begin{equation} \label{eq:reformulation2}
    \partial_t m_{t\to r}(x)
+
\partial_4 F\!\big[m_{t\to t}(x), x, t, t\big]\,
\partial_x m_{t\to r}(x).
\end{equation}
See Supplement~A.4 for a proof.
\end{theorem}
\revise{Here, the properties $G[f_3,f_4]|_{f_3=f_4}=1$ and $H[f_3,f_4]|_{f_3=f_4}=0$ are consistent with \autoref{thm:limitation} and ensure instantaneous consistency: when $f_3=f_4$, the right-hand side of \autoref{eq:reformulation} reduces to the instantaneous state $m_{t\to t}(x)$. The derivative structure in \autoref{eq:reformulation2} enables practical discretization of the derivative terms in the learning objective; see the numerical discussion in \autoref{sec:alg}.}



\begin{figure*}[t]
    \includegraphics[width=1.0\textwidth]{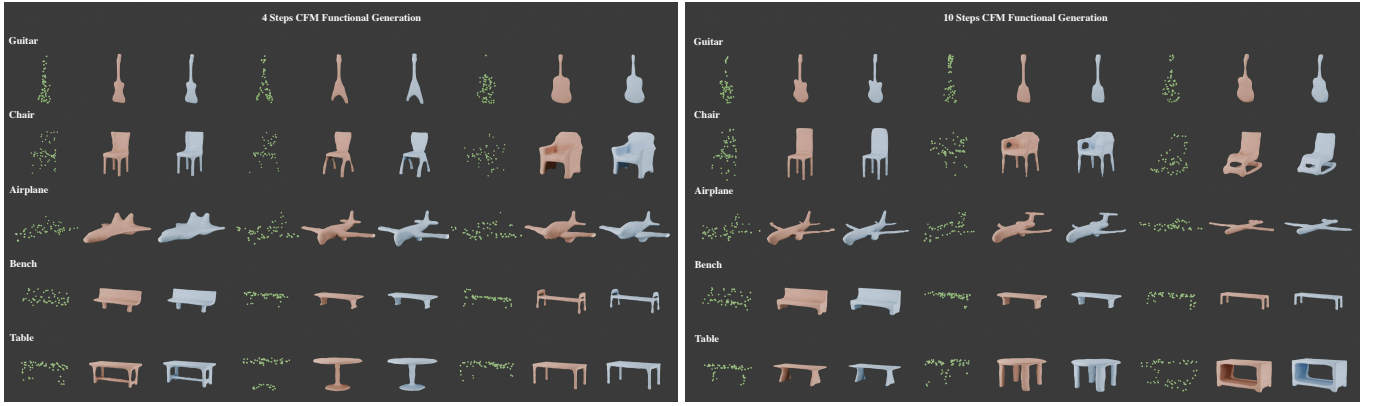}
    \caption{\revise{Few-step functional SDF generation from only 64 surface-conditioning points. We visualize results with 4 and 10 sampling steps, showing that our approach enables efficient generation by changing only the training objective, while leaving the model architecture unchanged and requiring no distillation.}}
    \label{fig:sdf_grid}
\end{figure*}

Based on this reformulation, the cumulative field loss $\mathcal{L}^{CMF}$ can be rewritten as $\|m^\theta_{t\to r}(x)-(G[t,r]m_{t\to t}(x)
+
H[t,r]E[m_{t\to r}(x),x,t,$\\$r\partial_t m_{t\to r}(x), \partial_x m_{t\to r}(x), m_{t\to t}(x)])\|_2^2$.  Since the resulting loss now involve the instantaneous field $m_{t\to t}(x)=m_t(x)$ and $E$ is affine to $m_{t\to t}(x)=m_t(x)$, we can replace it with the conditional instantaneous field $m_t(x | X_{t_1})$ to introduce supervision from the dataset.  In practice, we further use the current model prediction with stop-gradient applied as an estimator for $m_{t\to r}(x)$ on the right-hand side of \autoref{eq:reformulation}, which leads to the following surrogate loss:
\begin{equation}\label{eq:meanfields_loss}
\begin{aligned}
\mathcal{L}_c^{CFM}(\theta)= &\mathbb E_{t,r,x\sim P_t(\cdot|x_{t_1}), x_{t_1}\sim P_{data}}
\|m^\theta_{t\to r}(x)
-\text{sg}(\\
&G[t,r]m_{t}(x|x_{t_1}) 
+ H[t,r]E[
m^\theta_{t\to r}(x),x,t,r,\\
&\revise{\partial_t m^\theta_{t\to r}(x), \partial_x m^\theta_{t\to r}(x),
m_{t}(x|x_{t_1})
])\|_2^2.}
\end{aligned}
\end{equation}

 \begin{theorem}[Equivalence of Conditional and Marginal Losses]\label{thm:equivalence}
    We have $\mathcal{L}_c^{CMF}(\theta) = \mathcal{L}^{CMF}(\theta) + C$ where $C$ is independent of the model parameters~$\theta$.  (see Appendix~A.5 for proof.)
\end{theorem}
\autoref{thm:equivalence} implies that the computable loss $\mathcal{L}_c^{CMF}(\theta)$ can be used as a surrogate objective to optimize the target loss $\mathcal{L}^{CMF}(\theta)$. According to \autoref{thm:reformulation}, we have $G[t,t]=1$ and $H[t,t]=0$. Together with \autoref{thm:limitation}, when $t=r$ the loss reduces to the multi-step generation model loss
$\mathcal{L}_c = \mathbb{E}_{t,,x\sim P_t(\cdot\mid x_{t_1}),x_{t_1}\sim P_{\text{data}}}|m^\theta_{t\to r}(x)-m_t(x|x_{t_1})|$, which helps stabilize training, as shown in \autoref{tab:image_generation}.

 By specializing the loss $\mathcal{L}^{CMF}_c$ to different generative frameworks, we obtain the following instances (see Supplement A.5 for derivation in details).  Notably, our abstract function-based formulation significantly simplifies the resulting derivations.
 \begin{enumerate}
    \item  \textbf{\revise{$u-$FM:}} $ \mathcal{L}_c^{FM}(\theta) = \mathbb{E}_{ t, r,x_0\sim p_0, x_1\sim p_1, x = tx_1+(1-t)x_0}\|u_{t\to r}(x)-\text{sg}\big((r-t)(\partial_t u_{t\to r}(x)+(x_1-x_0)\partial_x u_{t\to r}(x))+(x_1-x_0)\big)\|_2^2$
    \item \textbf{\revise{$x_1-$FM:}} $\mathcal{L}_c^{FM}(\theta) = \mathbb{E}_{ t, r,x_0\sim p_0, x_1\sim p_1, x = tx_1+(1-t)x_0}\|x^1_{t\to r}(x)-\\\text{sg}\big(\frac{r-t}{1-r}((1-t)\partial_t u_{t\to r}(x)+(x_1-x)\partial_x u_{t\to r}(x))+x_1\big)\|_2^2$
    \item \textbf{\revise{DDIM:}} $\mathcal{L}_c^{FM}(\theta) \!= \!\mathbb{E}_{t, r,x_0\sim p_0, x_1\sim p_1, x = \sqrt{\bar \alpha_{t}}x_1+\sqrt{1-\bar \alpha_{t}}x_0} \|\tilde x_{t\to r}(x) -\\\text{sg}\big((\frac{\sqrt{1-\bar\alpha_{t}}\sqrt{\bar\alpha_{r}}}{\sqrt{1-\bar\alpha_{r}}\sqrt{\bar\alpha_{t}}}\!-\! 1)((\sqrt{\bar\alpha_{t}}x_1 \!-\!\bar\alpha_{t} x)\partial_x x_{0,t\to r}(x) -\frac{2(1-\bar\alpha_{t})(1-\beta_{t})}{\beta_{t}}\cdot\\\revise{\partial_{t} x_{0,{t}\to {r}}(x)) + x_1\big)\|_2^2}$
     \item \textbf{\revise{EDM:}} $\mathcal{L}_c^{FM}(\theta) = \mathbb{E}_{ t, r,x_{\sigma_{max}}\sim p_{\sigma_{max}}, x_0\sim p_0, x = x_0+tx_{\sigma_{max}}}\|D_{t\to r}(x)-\text{sg}(x_0+\frac{r-t}{r}(t\partial_t D_{t\to r}(x)+\partial_x D_{t\to r}(x)(x-x_0))\|_2^2$
 \end{enumerate}
The $u$-FM instantiation is mathematically equivalent to the Mean Flow, and $\text{sg}(\cdot)$ denote the stop gradient operation.\footnote{Our method is not a simple reparameterization of $x_1$-, $x_0$-, or $u$-prediction within the original MeanFlow framework \cite{geng2025mean}. Prior work \cite{li2025back} has shown that different prediction targets and loss formulations lead to fundamentally different behaviors, a conclusion that is also supported by our results in \autoref{sec:image_validation} and \ref{sec:geometry_distribution}. Moreover, reparameterization alone cannot capture more general formulations, such as $\psi_{t\to t+h}(x) = \sqrt{\bar\alpha_{t+h}}\tilde x_{t}(x_t) + \sqrt{1-\bar\alpha_{t+h}}\frac{x_t - \sqrt{\bar\alpha_t}\tilde x_{t}(x_t)}{\sqrt{1-\bar\alpha_t}}$ for DDIM. }

\section{Algorithms}\label{sec:alg}
Based on the above discussion, we obtain the training \revise{Algorithm} \autoref{alg:training} and the sampling \revise{Algorithm} \autoref{alg:sampling} for Cumulative Flow Map method.

\begin{algorithm}[t]
\caption{Cumulative Flow Map: Training}
\label{alg:training}
\begin{algorithmic}[1]  
\Require dataset $\mathcal{D}$, initial model parameter $\theta$, learning rate $\eta$, Normal Distribution $\mathcal{N}$, time sampler $\mathcal{T}$
\Repeat
    \State Sample $x_{t_0} \sim \mathcal{N}$ and $x_{t_1} \sim \mathcal{D}$   
    \State Sample $t, r \sim \mathcal{T}$ 
    \State Compute conditional distribution sample $x\sim p_t(x \mid X_{t_1} = x_{t_1})$ based on $x_{t_0}$, $x_{t_1}$ and $t$.  \hfill $\triangleright$ sec. \ref{sec:fundamental}
    \State Compute $\mathcal{L}_c^{CMF}(\theta)$  \hfill $\triangleright$ Eq.~\ref{eq:meanfields_loss} with instantiations (1)-(4)
    \State $\theta \gets \theta - \eta \nabla_\theta \mathcal{L}_c^{CMF}(\theta)$
\Until{convergence}
\end{algorithmic}
\end{algorithm}

\begin{algorithm}[t]
\caption{Cumulative Flow Map: Sampling}
\label{alg:sampling}
\begin{algorithmic}[1]  
\Require trained model parameter $\theta$, Normal Distribution $\mathcal{N}$, sampling steps $n$ 
\State Sample $x_{t_0} \sim \mathcal{N}$ 
\State Calculate sampling steps $\{\Delta t_k\}_{i=0}^{n-1}$ and $S_k=\sum_{i=0}^{k-1}\Delta t_k$  \hfill $\triangleright$ sec. \ref{sec:alg}
\If{$n==1$}
    \State $x_{t_1} \leftarrow F[m_{t_0\to t_1}(x_{t_0}),x_{t_0},t_0,t_1]$
\Else
    \For{$k = 0$ \textbf{to} $n-1$}
        \State $x_{S_k+\Delta t_k} \leftarrow F[m_{{S_k}\to {S_k+\Delta t_k}}^\theta(x_{S_k}), x_{S_k}, S_k, S_k+\Delta t_k]$
    \EndFor
\EndIf
\end{algorithmic}
\end{algorithm}

\paragraph{Time Sampler} During training, following \cite{geng2025mean}, we independently sample $t$ and $r$ from a distribution $\mathcal{T}_1$ and swap them if $r$ is closer to $t_0$ than $t$, forming the time sampler $\mathcal{T}$; for simplicity, we use $\mathcal{T}_1=\mathcal{U}[0,1]$ by default.  In addition, a fraction $\alpha$ of samples are constructed by setting $r = t$, which corresponds to training the instantaneous model $m_t^\theta$ when $t = r$ (as shown in \autoref{thm:limitation} and \autoref{thm:reformulation}).  Mixing a fraction of training $m_{t\to t}^\theta$ when training $m_{t\to r}^\theta$ improves training stability (see \autoref{tab:ablation_50k_multi}).  We set $\alpha = 0.5$ by default. During sampling, we use uniform time steps
$\Delta t_k = t_0 + k\,\frac{t_1 - t_0}{n}$ by default. 


\begin{figure*}[t]
\includegraphics[width=1.0\textwidth]{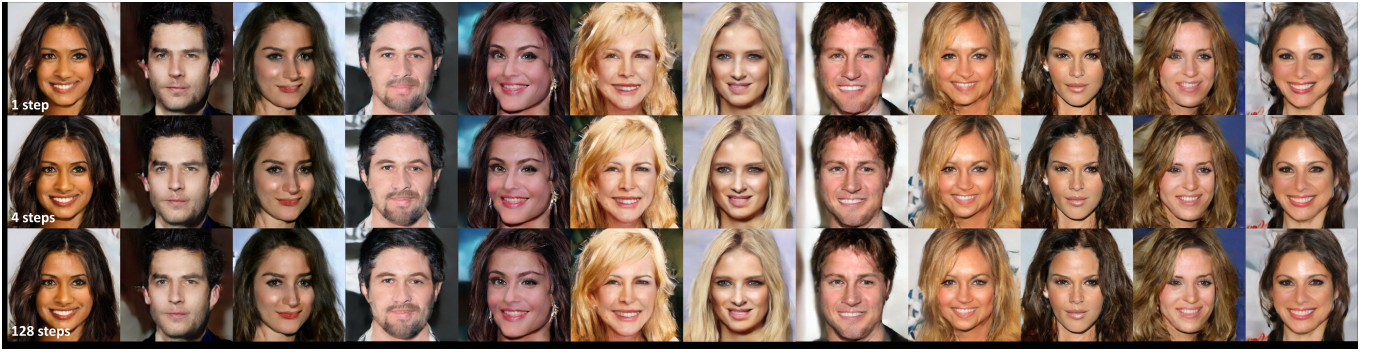}
\caption{\revise{Unconditional image generation results on the CelebA-HQ dataset using our CFM-DDIM training scheme. The resulting model supports efficient few-step sampling and produces visually comparable results with 1-step, 4-step, and 128-step generation, while achieving up to a \(128\times\) speedup. This improvement is obtained solely by modifying the training objective, without changing the model architecture or using distillation.}}
\label{fig:celeba}
\end{figure*}

\paragraph{Model Details and Training Acceleration}  CFM requires only minor modifications to the existing multi-step generative model architecture.  Specifically, we augment the original time embedder for $t$ with an additional embedder for $r$, and replace the original $\text{emb}_t$ with the averaged embedding $(\text{emb}_t + \text{emb}_r)/2$, where $\text{emb}_t$ and $\text{emb}_r$ denote the embeddings of $t$ and $r$ from the embedders, respectively.  By default, we adopt sinusoidal positional encoding for both embeddings.  Notably, under this design, when $r = t$, the averaged embedding reduces to $(\text{emb}_t + \text{emb}_r)/2 = \text{emb}_t$, making the model identical to the original multi-step formulation.  As a result, beyond training from scratch, CFM also support training few-step or one-step models from an existing multi-step model, initializing the $r$-embedder with parameters from $t$-embedder, which accelerates the training of one-step generation as shown in \autoref{sec:image_validation}.

\begin{figure*}[t]
    \includegraphics[width=1.0\textwidth]{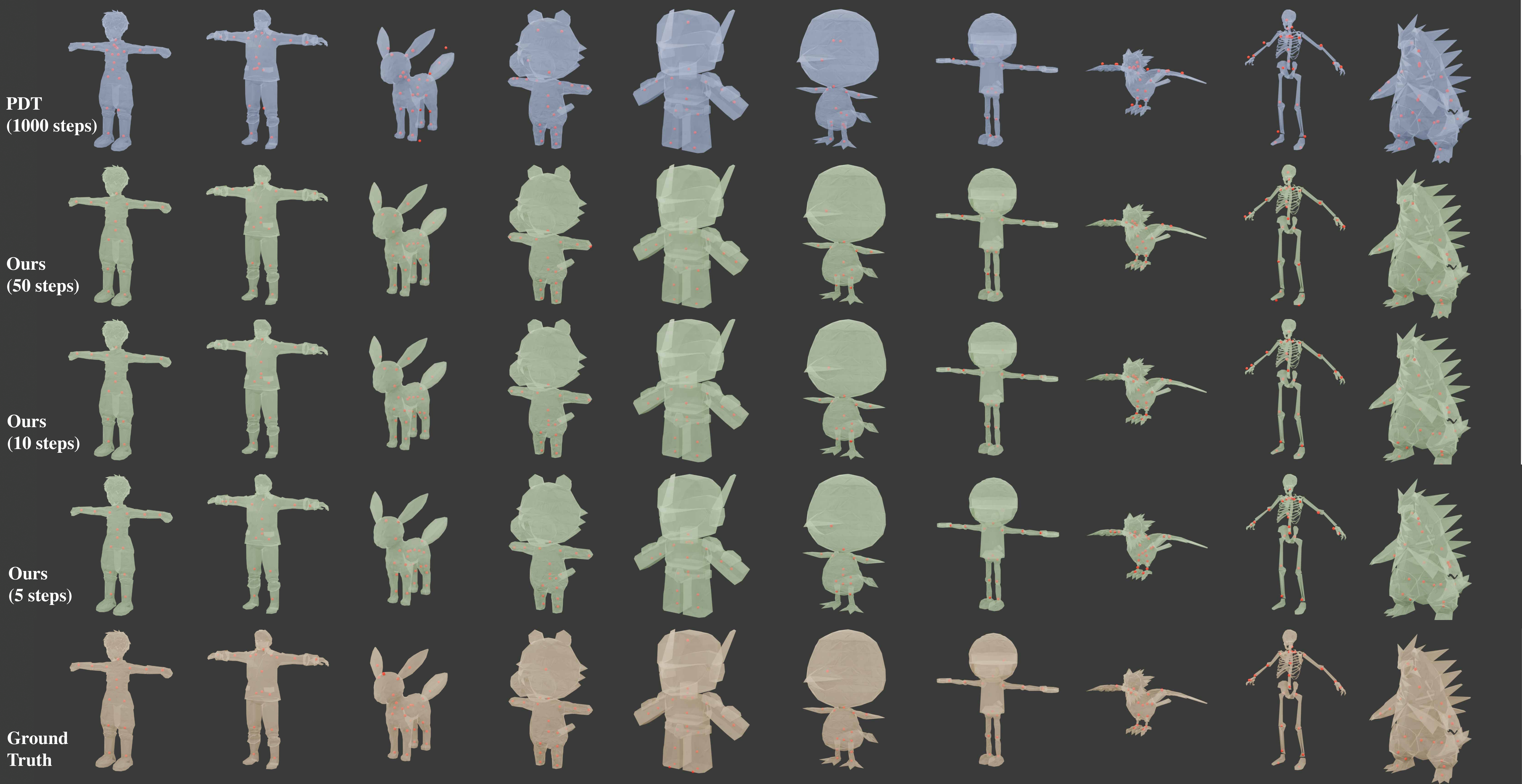}
    \vspace{-5mm}
    \caption{\revise{Joint generation with CFM-DDIM. Compared with the original PDT method, our approach delivers up to a \(200\times\) acceleration by only reformulating the training loss, without introducing architectural changes or relying on distillation.}}
    \label{fig:pdt_grid}
    \vspace{-3mm}
\end{figure*}

\paragraph{Gradient Calculation}
The loss in \autoref{eq:meanfields_loss} requires computing the derivatives of the model output $m_{t\to r}^\theta(x)$ with respect to both $t$ and $x$.
We consider two approaches. The first computes these derivatives using the Jacobian-vector product (JVP) operation provided by automatic differentiation frameworks such as PyTorch.  The second adopts a discrete approximation, which is applicable to neural
networks that do not support JVP.
The discrete method leverages the fact that
$\partial_t m_{t\to r}(x)$ and $\partial_x m_{t\to r}(x)$ appear jointly as shown in \autoref{thm:reformulation}, allowing their combination to be estimated as $\partial_t m_{t\to r}(x)+\partial_4 F\!\big[m_{t\to t}(x), x, t, t\big]\,
\partial_x m_{t\to r}(x)\approx \frac{m_{s\to r}(x+F[m_{t\to t}(x),x,t,t]h)-m_{t\to r}(x)}{h}= \frac{m_{s\to r}(\psi_{t\to t+h}(x))-m_{t\to r}(x)}{h}$, where $s = t + h$ denotes a time point close to $t$. In our experiments in \autoref{sec:point_cloud},\autoref{sec:sketch} and \autoref{sec:3D_SDF}, we use the JVP-based computation, and additionally validate the discrete approximation in \autoref{sec:image_validation} and \autoref{sec:geometry_distribution}.



\section{Experiments}\label{sec:experiment}
In this section, we evaluate our method on five graphics tasks, demonstrating that, using our approach, few-step generation can be achieved with only a minor modification to the model’s time embedding and the training loss, without additional architectural components or distillation procedures, thereby substantially accelerating generation while maintaining strong quality.  \textbf{Importantly, no single instantiation is optimal for all tasks}: for example, we show that EDM is necessary for Geometry Distribution, whereas only $x_1$-prediction flow matching supports few-step pixel-space image generation, with $u$-prediction methods failing in this setting.
\vspace{-3mm}
\subsection{Image Generation}\label{sec:image_validation}

\begin{table}[t]
\centering
\caption{
\revise{Image generation results on CelebA-HQ-256.
(a) Comparison of one-step and few-step generation capabilities of latent-space diffusion models trained for 400K steps.
FID-50k scores (lower is better) are reported for 128-, 4-, and 1-step denoising after training for 400K steps.
(b) Effectiveness of different training strategies evaluated at 100K training steps (see discussion in \autoref{sec:image_validation}).}}
\vspace{-3mm}
\label{tab:image_generation}

\begin{subtable}[t]{\linewidth}
\centering
\caption{Comparison}
\vspace{-2mm}
\label{tab:celeb_one_few_step}
\begin{tabular}{lccc}
\toprule
Method & 128-Step & 4-Step & 1-Step \\
\midrule
DDIM                     & 23.0 & 123.4 & 132.2 \\
Consistency Distillation & 59.5 & 39.6  & 38.2  \\
Consistency Training     & 53.7 & 19.0  & 33.2  \\
CFM-DDIM (Ours)          & \textbf{19.2} & \textbf{17.5} & \textbf{24.9} \\
\bottomrule
\end{tabular}
\end{subtable}

\vspace{1mm}

\begin{subtable}[t]{\linewidth}
\centering
\caption{Training strategy ablation}
\vspace{-2mm}
\label{tab:ablation_50k_multi}
\begin{tabular}{lccc}
\toprule
Setting & 128-Step & 4-Step & 1-Step \\
\midrule
Scratch                  & 27.0 & 36.6 & 46.9 \\
Self-distillation        & \textbf{17.4} & 35.1 & 42.7 \\
w/o mixing instantaneous & 572.3 & 572.3 & 572.3 \\
\bottomrule
\end{tabular}
\end{subtable}

\vspace{-6mm}
\end{table}

We first evaluate our method on the unconditional image generation task. We train few-step DDIM models from scratch on the CelebA-HQ dataset~\cite{karras2017progressive} using Algorithm~\autoref{alg:training} and the DDIM instantiation of the loss in Eq.~\ref{eq:meanfields_loss}, \revise{using a batch size of 64}. Following the latent-space generation paradigm, we adopt a DiT-B/2 backbone \cite{peebles2023scalable} with a standard sd-vae-ft-mse VAE \cite{rombach2022high}. We compare our approach with multi-step DDIM~\cite{song2020denoising} and prior diffusion-based few-step methods, including Consistency Models~\cite{song2023consistency} (both Consistency Training and Distillation).  As shown in \revise{Table~\ref{tab:celeb_one_few_step}}, our method achieves the best generation quality among the diffusion-related few-step generation methods considered here. \revise{We restrict this comparison to diffusion-based acceleration methods and exclude flow-matching-based methods such as Shortcut~\cite{frans2025shortcut}, since this experiment focuses on accelerating diffusion-based sampling.}\revise{\autoref{fig:celeba}} further shows that our approach attains comparable visual results with 1-step and 4-step sampling as with 128-step sampling.


In~\autoref{tab:ablation_50k_multi}, we analyze our training strategies in \autoref{sec:alg}. Scratch denotes training from random initialization, Self-Distillation continues training from a multi-step DDIM model pretrained for 50K steps (using the strategy described in the Model Details and Training Acceleration paragraph of~\autoref{sec:alg}), and w/o Mixing Instantaneous disables the mixing of instantaneous velocity during training (i.e., setting $\alpha=0$ for the $t=r$ case). The results show that few-step generation benefits from pretrained multi-step models and that mixing instantaneous velocity is essential, empirically supporting the applicability of \autoref{thm:limitation} and \autoref{thm:reformulation}.

\revise{Supplementary B.1} reports pixel-space image generation experiments on CelebA-HQ using the Just image Transformers (JiT) framework. We compare our $x_1$-FM few-step method with the $u$-prediction–based few-step method in MeanFlow~\cite{geng2025mean}, which can be viewed as a special case of our framework under $u$-FM. The results show that $u$-FM fails to support one-step generation in pixel space, highlighting the necessity of our method for flow-matching–related few-step generation.  

\subsection{Geometry Distribution}\label{sec:geometry_distribution}

The Geometry Distribution (GeoDist) task~\cite{zhang2025geometry} represents 3D shapes as point-cloud distributions $\mathcal{Q}$ and trains a generative network $D_\theta(P)$, $P=(x,y,z)$, to transform noise samples into point clouds drawn from $\mathcal{Q}$, thereby compressing geometry into the network parameters $\theta$. GeoDist adopts the EDM formulation for generation. We sample $n=2^{25}$ points from shape surfaces to form the training set $\{Q_i\}_{i=1}^n$, and train few-step EDM models using Algorithm~\autoref{alg:training} and the EDM instantiation of the loss in Eq.~\ref{eq:meanfields_loss}. Following~\cite{zhang2025geometry}, we use an MLP with a matched number of parameters and 3D inputs and outputs, evaluate a diverse set of shapes with complex topology, thin structures, and complex scenes, \revise{and use Chamfer Distance to measure the similarity between point clouds generated by the network and point samples from the ground-truth shapes. As shown in \autoref{tab:geometry_distribution_compare}, CFM-EDM is the most effective choice for few-step generation on this task, showing stronger robustness, with no significant differences in Chamfer Distance observed among 3-step, 6-step, and 60-step sampling, whereas the $x_1$-FM variant demonstrates a certain degree of few-step generation capability, and the original model and Mean Flow do not support few-step sampling for this task. As shown in \autoref{fig:geodist_grid}, compared to the original 60-step generation, our method achieves comparable reconstruction quality with 6$\times$ and 10$\times$ speedups.}

\begin{table}[H]
\vspace{-5mm}
\centering
\caption{\revise{Comparison of few-step generation performance on the Geometry Distribution task (Chamfer Distance, lower is better).}}
\label{tab:geometry_distribution_compare}
\small
\setlength{\tabcolsep}{5pt}
\begin{tabular}{lccc}
\hline
\textbf{Method} & 60-step & 6-step & 3-step  \\
\hline
GeoDist  &0.017  &0.119  &50.153  \\
\hline
$x_1-$pred FM (Ours)  &0.017  &0.031  &0.064  \\
\hline
$u-$pred FM (Ours)  &0.630  &0.629  &0.628  \\
\hline
EDM (Ours)  &0.017  &0.018  &0.018  \\
\hline
\end{tabular}
\vspace{-5mm}
\end{table}

\begin{table*}[t]
\centering
\caption{
\revise{Quantitative evaluation across multiple tasks.
(a) PDT: Metrics of joint prediction results for point distribution transformation (PDT). 
PDT results are computed using publicly released checkpoints. 
Our method enables few-step inference while maintaining generation quality.
(b) SDF: Quantitative comparison of reconstruction quality on the ShapeNet dataset. 
The model is trained with a conditional input of 64 points sampled from the target surface and is required to reconstruct the surface from these points;
(c) Sketch: Quantitative comparison of image-sketch fidelity on the ControlSketch dataset
}}
\vspace{-3mm}
\label{tab:quantitative_results}

\begin{minipage}[t]{0.44\linewidth}
\centering
\begin{subtable}[t]{\linewidth}
\centering
\caption{PDT}
\vspace{-2mm}
\label{tab:pdt}
\begin{tabular}{lcccc}
\toprule
Method & CD-J2J($\downarrow$) & IoU($\uparrow$) & Prec.($\uparrow$) & Rec.($\uparrow$)\\
\midrule
PDT (DDPM 1000-step)  & 6.4\% & 57.4\% & \underline{53.6}\% & 64.5\% \\
PDT (DDPM 50-step)   & 26.6\% & 1.0\%  & 0.5\%  & 42.7\% \\
PDT (DDPM 10-step)   & 27.8\% & 0.8\%  & 0.4\%  & 36.3\% \\
Ours (CFM-DDIM 50-step) & \underline{5.4\%} & \textbf{66.9}\% & \textbf{60.8\%} & \textbf{77.7\%} \\
Ours (CFM-DDIM 10-step) & \textbf{5.2\%} & \underline{66.3}\% & \textbf{60.8}\% & \underline{76.9}\% \\
Ours (CFM-DDIM 5-step)  & 6.2\% & 54.3\% & 47.3\% & 67.8\% \\
\bottomrule
\end{tabular}
\end{subtable}

\begin{subtable}[t]{\linewidth}
\centering
\caption{SDF}
\vspace{-2mm}
\label{tab:reconstruction}
\begin{tabular}{lccc}
\toprule
Method &  Chamfer $\downarrow$ & F-Score $\uparrow$ & Boundary $\downarrow$ \\
\midrule
Ours (4-step)  & 0.048 & 0.659 & 0.011 \\
Ours (10-step) & 0.048 & 0.660 & 0.011 \\
FD (64-step)   & 0.101 & 0.707 & 0.012 \\
\bottomrule
\end{tabular}
\end{subtable}
\end{minipage}
\hfill
\begin{minipage}[t]{0.52\linewidth}
\centering
\begin{subtable}[t]{\linewidth}
\centering
\caption{Sketch}
\vspace{-2mm}
\label{tab:sketch}
\begin{tabular}{lcccc}
\toprule
 & \multicolumn{2}{c}{MS-SSIM $\uparrow$} & \multicolumn{2}{c}{DreamSim $\downarrow$} \\
\cmidrule(lr){2-3} \cmidrule(lr){4-5}
 & Seen & Unseen & Seen & Unseen \\
\midrule
Cat (SwiftSketch 50-step) & 0.619 & 0.614 & 0.577 & 0.577 \\
Cat (CFM 4-step)         & 0.618 & 0.612 & 0.578 & 0.577 \\
Cat (CFM 1-step)         & 0.617 & 0.611 & 0.579 & 0.576 \\
\midrule
Fish (SwiftSketch 50-step) & 0.589 & 0.590 & 0.567 & 0.570 \\
Fish (CFM 4-step)          & 0.589 & 0.590 & 0.568 & 0.570 \\
Fish (CFM 1-step)          & 0.589 & 0.590 & 0.569 & 0.571 \\
\midrule
Rabbit (SwiftSketch 50-step) & 0.691 & 0.691 & 0.538 & 0.542 \\
Rabbit (CFM 4-step)          & 0.690 & 0.691 & 0.537 & 0.542 \\
Rabbit (CFM 1-step)          & 0.688 & 0.688 & 0.538 & 0.543 \\
\bottomrule
\end{tabular}
\end{subtable}

\end{minipage}

\end{table*}

\subsection{PDT}\label{sec:point_cloud}
\begin{figure}[t]
    \centering
    \includegraphics[width=\linewidth]{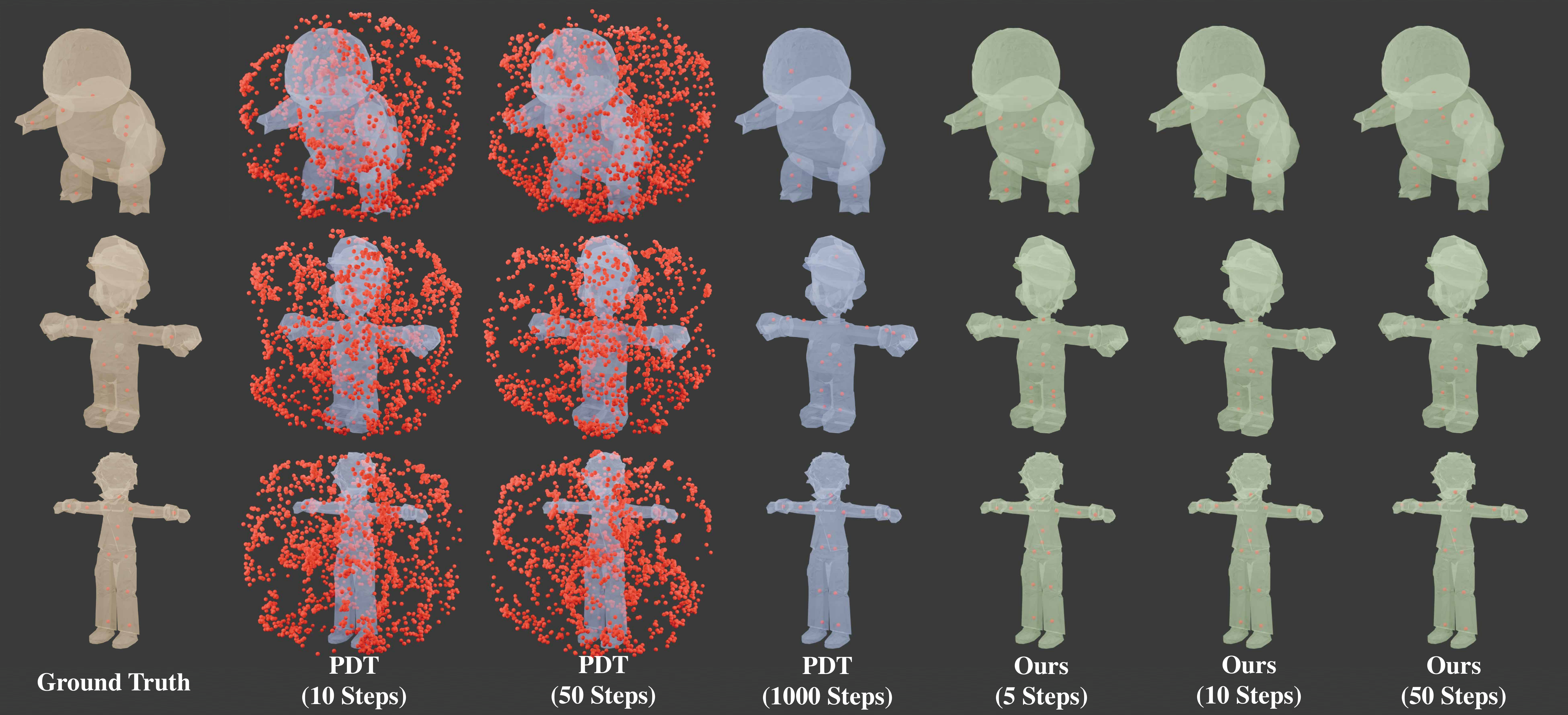}
    \caption{Comparison between original PDT method using 10 inference steps, 50 inference steps, 1000 inference steps and our CFM-DDIM method using 5 inference steps ($200\times$ speedup), \revise{10} inference steps ($100\times$ speedup) and 50 inference steps ($20\times$ speedup).}
    \label{fig:pdt_comparison}
    \vspace{-7.5mm}
\end{figure}

We evaluate our CFM method on the joint position prediction task using the RigNet dataset~\cite{RigNet}, and compare against Point Distribution Transformation (PDT)~\cite{wang2025pdt}. PDT learns a conditional transformation that maps an input point cloud from its original geometric distribution to a target distribution corresponding to joint locations. While PDT is originally trained and evaluated with DDPM sampling using 1000 inference steps, we adapt it to DDIM sampling and incorporate our Cumulative Field modification, enabling up to a 200$\times$ reduction in sampling cost (5 inference steps) while preserving prediction quality relative to the original PDT. For a controlled comparison, we retain the PDT architecture, using PVCNN~\cite{liu2019point} to extract features from the conditioning point cloud, followed by eight DiT-3D \cite{mo2023dit} blocks for joint generation.
Table~\ref{tab:pdt} reports a quantitative comparison with PDT under varying numbers of inference steps. We follow the evaluation introduced in~\cite{RigNet}: CD-J2J measures the mean bidirectional nearest-neighbor distance between predicted and reference joints, while IoU, Precision, and Recall are computed via Hungarian matching, capturing the fraction of mutually matched joints, the fraction of predicted joints matched within tolerance, and the fraction of reference joints matched within tolerance, respectively. Qualitative comparisons are provided in Fig.~\ref{fig:pdt_grid} and validation experiment on using fewer inference steps for original PDT are provided in Fig.~\ref{fig:pdt_comparison}.

\subsection{Image-Based Sketch Generation}\label{sec:sketch}
Given an input image $I$, we aim to generate a sketch $S$ that faithfully reflects the input while retaining a natural sketch-like appearance. The sketch consists of multiple strokes, each represented as a B\'ezier curve defined by control points. \citet{arar2025swift} address this task using a conditional diffusion model. Starting from randomly initialized point sets, they train a Transformer decoder on ControlSketch dataset \cite{arar2025swift} to iteratively denoise the points and produce vectorized sketches. The decoder incorporates cross-attention with image features extracted by a pretrained CLIP image encoder \cite{radford2021clip}, enabling effective conditioning on the input image.

We evaluate the performance of CFM on the ControlSketch dataset and compare it against SwiftSketch \cite{arar2025swift}. While SwiftSketch requires 50 diffusion steps during inference, our method generates results in 4 steps or only 1 step, achieving up to 50× inference speedup. We conduct experiments on three categories from the ControlSketch dataset. To ensure a fair comparison, we adopt the same model architecture as SwiftSketch and train both SwiftSketch and CFM separately on each category for 50,000 steps. Following SwiftSketch, we apply a refinement stage at test time, after which MS-SSIM \cite{wang2003multi} and DreamSim \cite{fu2023dreamsim} scores are computed to quantitatively evaluate image–sketch fidelity. We report MS-SSIM and DreamSim scores on both the training set (seen) and the validation set (unseen) of ControlSketch dataset in \autoref{tab:sketch}. The results show that CFM achieves performance comparable to SwiftSketch across both metrics on seen and unseen images.

\begin{figure}[t]
    \centering
    \includegraphics[width=\linewidth]{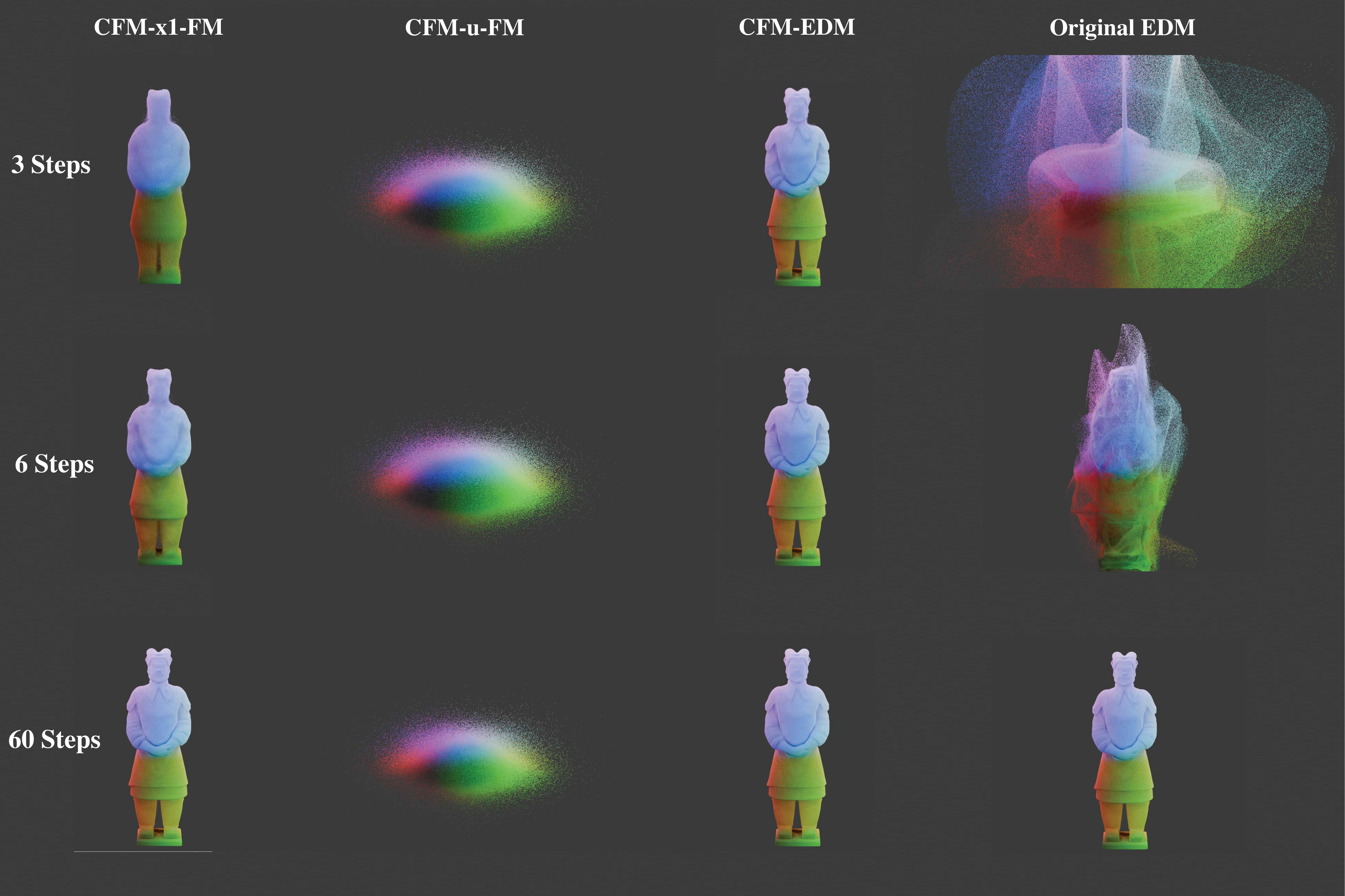}
    \caption{We show that CFM-EDM works best under the application of geometric distribution comparing to CFM $x_1$-FM and CFM $u$-FM.}
    \label{fig:geodist_ab}
    \vspace{-5mm}
\end{figure}

\subsection{3D SDF Generation}\label{sec:3D_SDF}

Functional Diffusion \cite{zhang2024functional} (FuncGen) introduces a challenging sparse conditional generation task: given only 64 surface points as conditions, the model reconstructs the full SDF of a shape. This task is effectively addressed only by Functional Diffusion through its function-based generative framework. We evaluate our $x_1$-FM few-step generation method on this task. We train the model using Algorithm~\autoref{alg:training} and the loss in \revise{Section~\ref{sec:training_meanfield}}, adopting the self-attention–based architecture proposed in \cite{zhang2024functional}. Both inputs and outputs are represented as functions via randomly sampled point–value pairs. Specifically, the input function $f_c$ is represented by context points and values \revise{$\{(x_c^i, v_c^i)\}_{i=1}^n$}, while the output function $f_q$ is represented by query points and predicted values \revise{$\{(x_q^j, v_q^j)\}_{j=1}^m$}. Functional Diffusion conditions on 64 surface points and reconstructs the target SDF through 64 denoising steps.

We evaluate the models using Chamfer Distance, F1-score, and Boundary Loss, following prior work~\cite{zhang2024functional,zhang20233dshape2vecset}. Chamfer Distance measures the bidirectional distance between generated and ground-truth surfaces, F1-score captures surface reconstruction accuracy, and Boundary Loss measures the mean squared error of predicted SDF values near the zero-level surface. Chamfer Distance and F1-score are computed using 50K uniformly sampled surface points, while Boundary Loss is evaluated on 100K near-surface samples. We use the same train/test split as~\cite{zhang2024functional}.  As shown in \autoref{tab:reconstruction}, our method achieves 6-16× speedups over Functional Diffusion while maintaining comparable reconstruction quality.




\section{Additional Experiments and Discussion}
\paragraph{Toy Example} \revise{To better visualize the sample positions and prediction targets at each step under different formulations, we conduct a 2D toy experiment using an MLP on the standard Checkerboard and Two-Moons datasets with 4-step sampling. At each step, we plot the sample positions $x_t$ and the corresponding predicted cumulative fields $m_{t\to r}(x_t)$. The resulting visualization in \autoref{fig:toy_example} highlights the differences among different formulations during few-step generation.}

\begin{figure}[t]
    \centering
    \vspace{-2mm}
    \begin{subfigure}{0.8\linewidth}
        \centering
        \includegraphics[width=\linewidth]{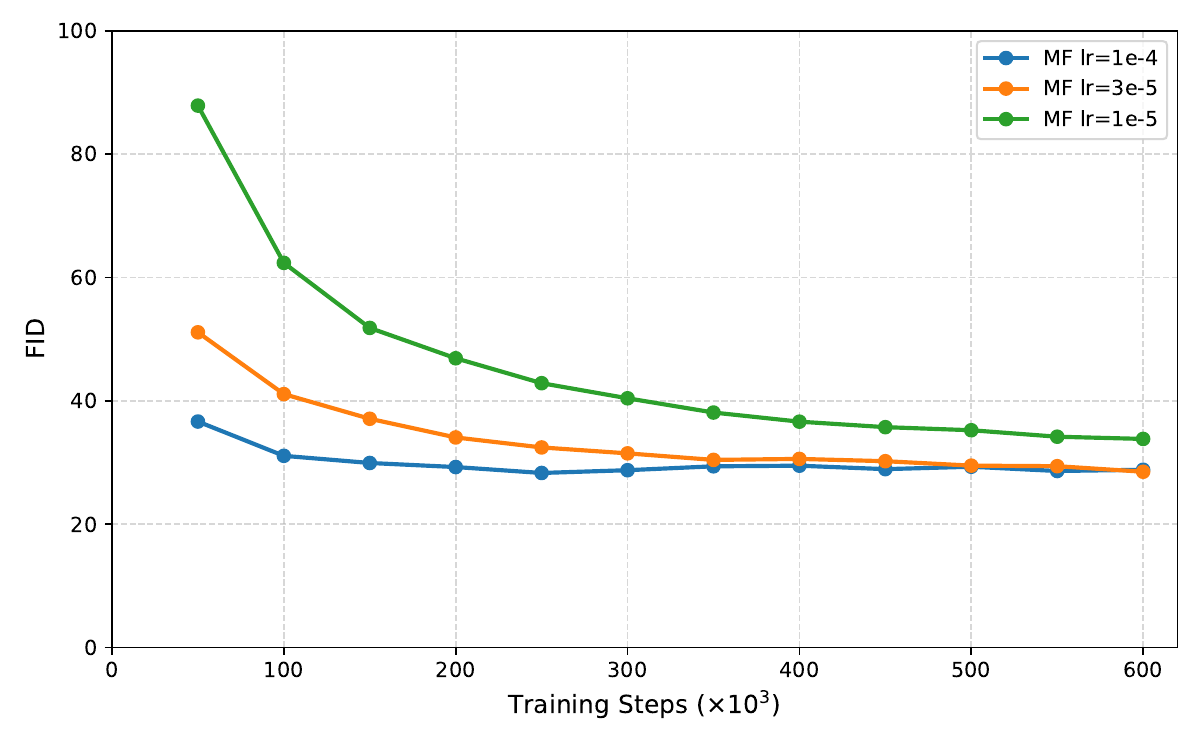}
        \vspace{-5mm}
        \caption{FID-50k score pf MeanFlow with 1-step generation}
        \vspace{-5mm}
    \end{subfigure}
    \vspace{0.5cm} 
    \begin{subfigure}{0.8\linewidth}
        \centering
        \includegraphics[width=\linewidth]{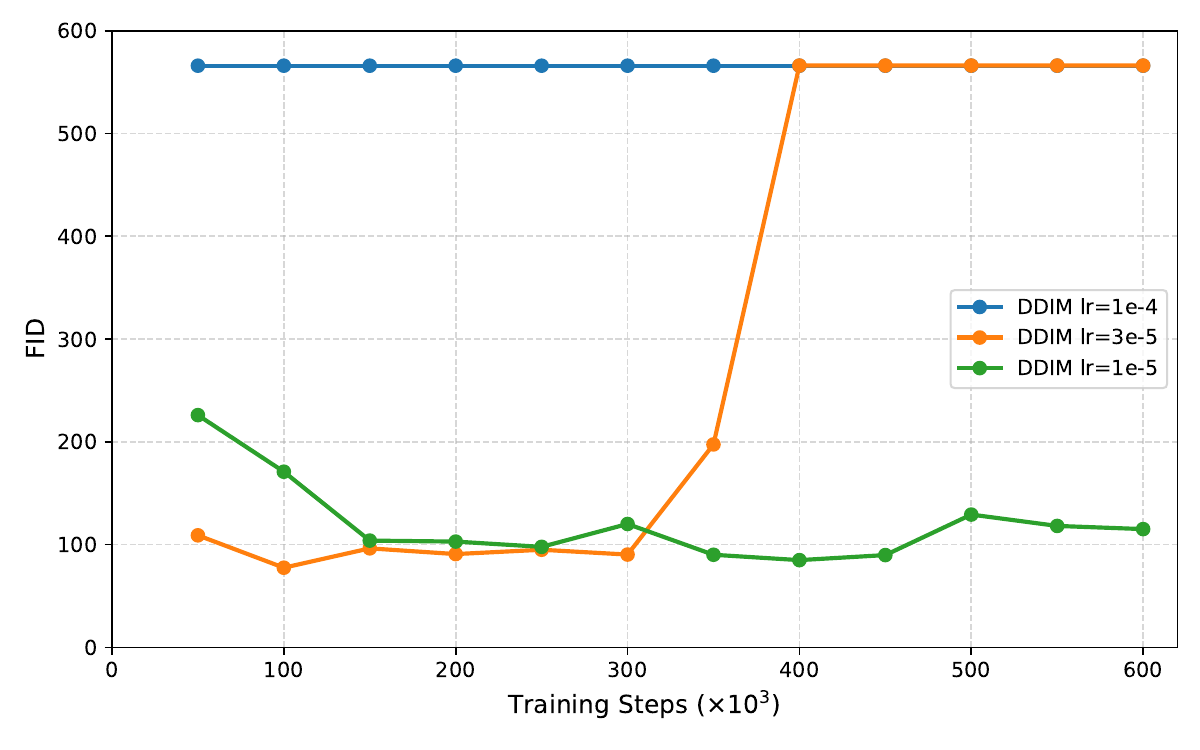}
        \vspace{-5mm}
        \caption{FID-50k score of DDIM with 1-step generation}
        \vspace{-9mm}
    \end{subfigure}
    \caption{Comparison of the learning rate sensitivity of MeanFlow and DDIM.}
    \label{fig:sensitivity}
    \vspace{-5mm}
\end{figure}

\paragraph{Parameter Study}\revise{In the main experiments, we compare the EDM, $x_1$-prediction FM, and $u$-prediction FM instantiations of CFM, as shown in \autoref{sec:image_validation} and \autoref{sec:geometry_distribution}. Here, we further compare the learning-rate sensitivity of the $u$-prediction FM and DDIM instantiations. Both models are trained on the CelebA-HQ dataset using a DiT-B/2 architecture and the same configuration as in \autoref{sec:image_validation}, except that we use a batch size of 32 for efficiency. During training, we measure the FID-50K score every 50K steps using one-step generation and report the results in \autoref{fig:sensitivity}.  The results indicate that DDIM is more sensitive to the learning rate. When trained for 400K iterations with a learning rate of $1\times10^{-5}$, DDIM achieves its best FID-50K score of 84.94, whereas training with a learning rate of $1\times10^{-4}$ keeps the FID-50K score high at 566.11 after 400K iterations. In contrast, $u$-prediction FM converges to comparable FID scores across learning rates of $1\times10^{-4}$, $3\times10^{-5}$, and $1\times10^{-5}$. For DDIM, when the learning rate is further reduced to $1\times10^{-6}$ after 400K iterations and training is continued, the FID-50K score improves to 65.89, further confirming its stronger sensitivity to the learning rate.
}

\paragraph{Discussion of Training Cost}\revise{Since CFM does not require changes to the model architecture, batch size, or other main training configurations, its additional training cost mainly comes from the computation of the loss in \autoref{eq:meanfields_loss}. This loss involves derivative terms, for which we provide two practical computation strategies in \autoref{sec:alg}: one based on Jacobian-vector products (JVPs) in automatic differentiation frameworks such as PyTorch, and the other based on finite differences. The JVP-based implementation requires additional forward-mode differentiation, while the finite-difference implementation requires extra model evaluations to approximate the derivative terms. Both choices therefore introduce additional computational overhead.  The exact overhead depends on the task, model architecture, and chosen derivative computation strategy. In our experiments, we observe that the training-time cost of CFM is typically about $2\times$--$3\times$ that of the corresponding multi-step training baseline. The overhead is smallest for image generation, at about $2\times$ the training cost, and largest for geometry distribution, at about $3\times$.}

\section{Conclusion}
We presented \emph{Cumulative Flow Maps (CFM)}, a unified framework for few-step generation (including one-step generation).  With only minor architectural modifications and changes to the training loss, CFM enables few-step inference for a broad class of multi-step generative models, including $u$- and $x_1$-prediction flow matching, DDIM, and EDM. Our approach generalizes beyond prior methods such as Mean Flow by enabling few-step generation on graphics tasks that are not supported by these methods, achieving substantial speedups (up to 10$\times$–200$\times$) while maintaining strong generation quality. While CFM is, in principle, applicable to a wide range of generative models and tasks, our evaluation is limited to five applications and four representative formulations.  \revise{Extending CFM to additional tasks, broader generative paradigms, and larger-scale datasets such as ImageNet remains an important direction for future work.}

\begin{acks}
We express our gratitude to the anonymous reviewers for their insightful feedback. Georgia Tech authors acknowledge NSF IIS \#2433322, ECCS \#2318814, CAREER \#2433307, IIS \#2106733, OISE \#2433313, and CNS \#1919647 for funding support.
\end{acks}

\bibliographystyle{ACM-Reference-Format}
\bibliography{refs_application.bib, refs_generative.bib, refs_flowmap.bib}

\newpage

\end{document}